\documentclass{article}

\usepackage{microtype}
\usepackage{graphicx}
\usepackage{subcaption}
\usepackage{booktabs} 

\usepackage{hyperref}

\usepackage[preprint]{icml2026}

\usepackage{amsmath}
\usepackage{amssymb}
\usepackage{mathtools}
\usepackage{amsthm}
\usepackage{xurl}
\usepackage{balance}

\usepackage[capitalize,noabbrev]{cleveref}

\theoremstyle{plain}

\theoremstyle{definition}

\theoremstyle{remark}

\usepackage[textsize=tiny]{todonotes}

\usepackage{amsmath,amsthm,amssymb}
\usepackage{cleveref}
\usepackage{url}

\usepackage{xspace}
\usepackage{dsfont}
\usepackage{afterpage}

\usepackage{enumitem}
\usepackage{booktabs}
\usepackage{caption} 
\usepackage{multirow} 
\usepackage{enumitem}
\usepackage{colortbl} 
\usepackage{bbm}

\usepackage{listings}
\usepackage{xcolor}
\usepackage{tcolorbox}
\tcbuselibrary{listings, skins, breakable, theorems}
\usepackage{lipsum} 
\usepackage{subcaption} 
\usepackage{mwe}

\usepackage{listings}
\usepackage{graphicx,color}
\definecolor{lightblue}{RGB}{220, 240, 255}
\newtcolorbox{mybluebox}[1]{%
  breakable,
  colback=lightblue,
  colframe=blue!60,
  fonttitle=\bfseries,
  title=#1,
  width=0.9\textwidth,
  left=5pt,
  right=5pt,
  boxrule=1pt
}
\definecolor{lightyellow}{RGB}{250, 240, 220}
\newtcolorbox{myyellowbox}[1]{%
  breakable,                
  colback=lightyellow,
  colframe=orange!40,
  fonttitle=\bfseries,
  title=#1,
  width=0.95\textwidth,
  left=5pt,
  right=5pt,
  boxrule=1pt
}
\usepackage{mathtools} 
\usepackage{bm} 
\usepackage{soul} 
\usepackage{nicefrac}
\usepackage{csquotes}

\usepackage{wrapfig}
\usepackage{duckuments}

\definecolor{cornellred}{rgb}{0.7, 0.11, 0.11}
\definecolor{cadmiumgreen}{rgb}{0.0, 0.42, 0.24}
\definecolor{aliceblue}{rgb}{0.91, 0.94, 0.97}
\definecolor{darkblue}{rgb}{0.83, 0.89, 0.97}
\definecolor{Red}{rgb}{0.941, 0.243, 0.243}
\definecolor{Green}{RGB}{55, 178, 77}
\definecolor{Blue}{rgb}{0.098,0.3,0.9}

\definecolor{codegreen}{rgb}{0.25,0.5,0.25}
\definecolor{codepurple}{rgb}{0.5,0,0.35}
\definecolor{codeblue}{rgb}{0.1,0.1,0.7}
\definecolor{backcolour}{rgb}{0.97,0.97,0.97}
\definecolor{framecolour}{rgb}{0.6,0.6,0.6}
\definecolor{titlebg}{rgb}{0.9,0.9,0.9}

\newtcblisting{pythonsnippet}[2][]{
  enhanced,
  title={#2},
  label={snippet:#2},
  colback=backcolour,
  colframe=framecolour,
  coltitle=black,
  colbacktitle=titlebg,
  fonttitle=\bfseries\small\ttfamily,
  attach boxed title to top left={xshift=10pt, yshift*=-\tcboxedtitleheight/2},
  boxed title style={boxrule=0.5pt, sharp corners=down, rounded corners=northeast, arc=3pt},
  listing only,
  listing options={
    language=Python,
    basicstyle=\ttfamily\scriptsize,
    keywordstyle=\color{codeblue}\bfseries,
    commentstyle=\color{codegreen}\itshape,
    stringstyle=\color{codepurple},
    numbers=left,
    numberstyle=\tiny\color{gray},
    stepnumber=1,
    numbersep=8pt,
    tabsize=4,
    showstringspaces=false,
    breaklines=true,
    breakatwhitespace=true,
    aboveskip=0pt,
    belowskip=0pt,
  }
}
  
\usepackage{pifont}

\sethlcolor{aliceblue}

\newcommand{\longname}{\textbf{M}eta-\textbf{A}wareness via \textbf{P}redictive \textbf{R}eward\xspace}
\newcommand{\metaname}{\textbf{MAPR}\xspace}
\newcommand{\effname}{\textbf{MAPR}\xspace-\textit{efficient}}
\newcommand*{\ShowNotes}{} 

\ifdefined\ShowNotes
  \newcommand{\colornote}[3]{{\color{#1}\bf{#2: #3}\normalfont}}
\else
  \newcommand{\colornote}[3]{}
\fi

\definecolor{darkred}{RGB}{177, 38, 26}
\definecolor{darkblue}{RGB}{67, 116, 177}
\definecolor{darkgreen}{rgb}{0.0, 0.5, 0.0}
\definecolor{bestcol}{RGB}{  0,102,204} %
\definecolor{goodcol}{RGB}{ 34,139, 34} %

\newcommand{\rowhighlight}{\rowcolor{aliceblue}}

\hypersetup{
  linkcolor = cornellred,
  citecolor  = cadmiumgreen,
  colorlinks = true,
  urlcolor = Blue
}

\usepackage{amsmath,amsfonts,bm}

\def\eqref#1{equation~\ref{#1}}

\def\1{\bm{1}}

\DeclareMathAlphabet{\mathsfit}{\encodingdefault}{\sfdefault}{m}{sl}
\SetMathAlphabet{\mathsfit}{bold}{\encodingdefault}{\sfdefault}{bx}{n}

\icmltitlerunning{Verifying Meta-Awareness via Predictive Rewards in Reasoning Models}

\begin{document}

\twocolumn[
  \icmltitle{Verifying Meta-Awareness via Predictive Rewards in Reasoning Models}
  \icmlsetsymbol{equal}{*}

  \begin{icmlauthorlist}
    \icmlauthor{Yoonjeon Kim}{equal,+}
    \icmlauthor{Doohyuk Jang}{equal,+}
    \icmlauthor{Eunho Yang}{+,@}
  \end{icmlauthorlist}

  \icmlaffiliation{+}{KAIST, Daejeon, South Korea}
  \icmlaffiliation{@}{AITRICS, Seoul, South Korea}

  \icmlcorrespondingauthor{Eunho Yang}{eunhoy@kaist.ac.kr}
  \icmlkeywords{Machine Learning, ICML}

  \vskip 0.3in
]

\printAffiliationsAndNotice{\icmlEqualContribution} 

\begin{abstract}
  Recent research on reasoning models explores the meta-awareness of language models, including their ability to determine optimal thinking duration, recognize knowledge boundaries, and structure concept-level thinking. While current large reasoning models depend solely on answer-based verification, we show that adding meta-awareness objectives leads to significant performance gains over models without such meta-knowledge. \metaname{} (\longname{}) utilizes a self-generated task of predicting rollout statistics - specifically length, pass-rate, and concepts used - allowing for verification against the actual statistics. Furthermore, by leveraging this self-predictive capability, the model can regulate its reasoning behavior by i) filtering out trivial or unsolvable prompts, ii) reducing lengthy generations that tend to be incorrect, and iii) generating hints relevant to the problem. The results are inspiring: \metaname{} yields significant improvements in both accuracy and training efficiency on various reasoning benchmarks. More specifically, our method can speed up GRPO training by over 1.28$\times$ to reach the same performance, and achieve 83.18\% gain in accuracy on AIME25, and a 13.04\% average gain over six mathematics benchmarks. The code is publicly available at \url{https://github.com/akatigre/MAPR-RL}.
\end{abstract}

\section{Introduction}
Recent studies have confirmed that applying RL-based post-training to large language models (LLMs) \citep{brown2020language, yang2025qwen3, touvron2023llama} can significantly enhance their reasoning ability. In particular, methods such as GRPO \citep{deepseekmath_grpo}, which efficiently train large reasoning models (LRMs) \citep{guo2025deepseek, acereason_nemotron} without an explicit critic model, have recently attracted considerable attention.

Beyond the success of LRMs, the paradigm of meta-awareness, which is the ability to recognize its own knowledge and ignorance, has drawn increasing attention from the research community \citep{sui2025meta, mera, de2024rational, chen2025aware, liu2025ghpo, zhang2025adaptthink, shen2025dast, tu2025learning, shi2025efficient,qu2025optimizing}. However, existing approaches remain constrained by their reliance on external model, curated dataset and reasoning pipelines that require human intervention.

\begin{figure}
    \centering
    \begin{subfigure}[t]{0.48\textwidth}
      \centering
      \includegraphics[width=\textwidth]{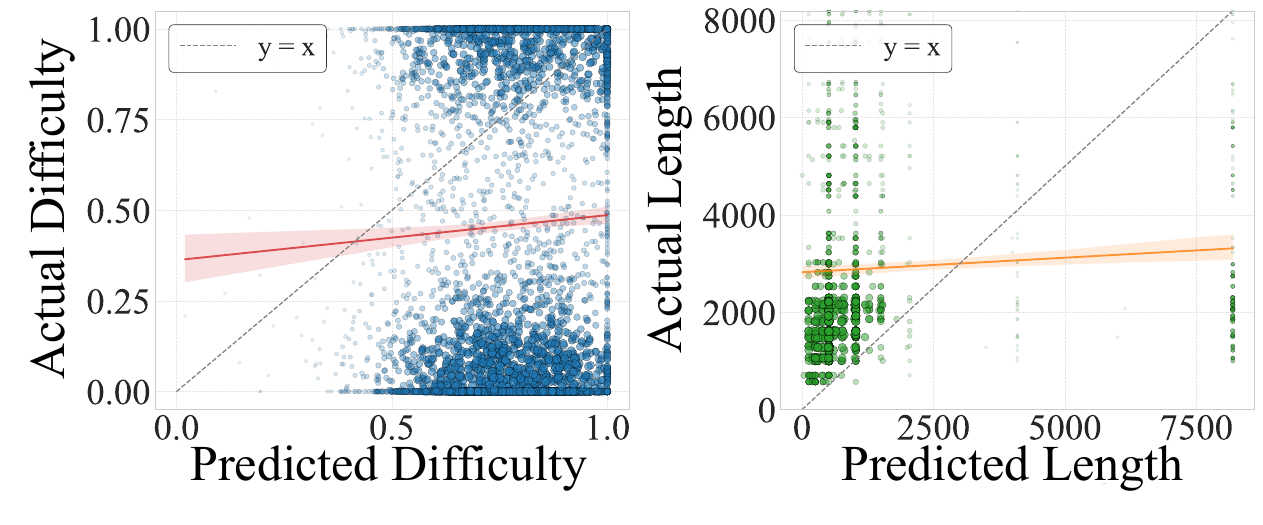}
      \vspace{-0.2in}
      \caption{Poor Alignment of GRPO Trained Model.}
      \label{fig:a1}
    \end{subfigure}
    \hfill
    \begin{subfigure}[t]{0.48\textwidth}
      \centering
      \includegraphics[width=\textwidth]{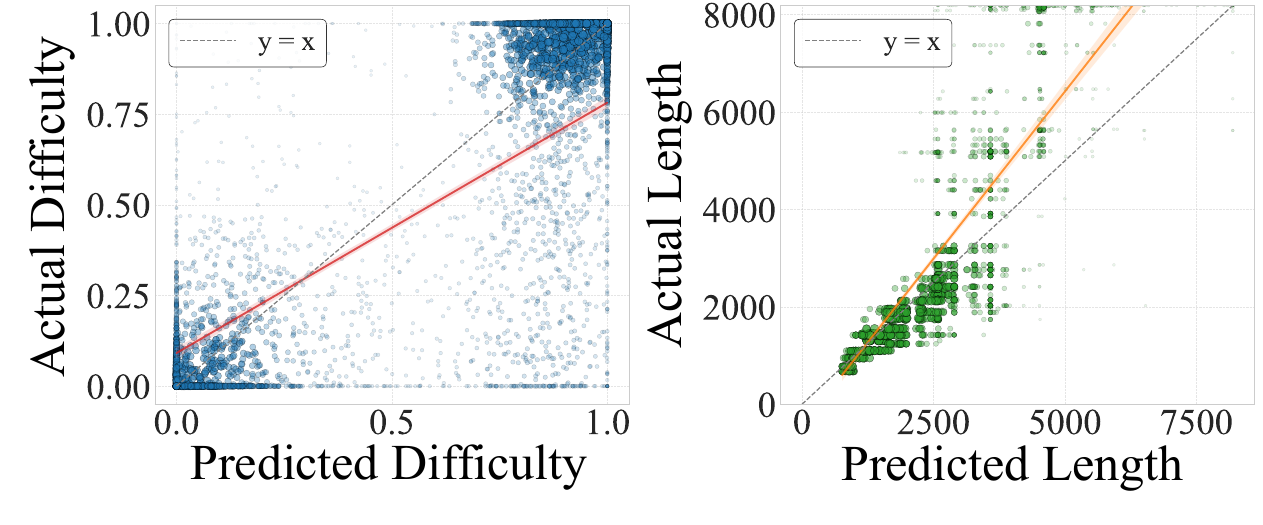}
      \vspace{-0.2in}
      \caption{Enhanced Alignment of \metaname{} Trained Model.}
      \label{fig:a2}
    \end{subfigure}
    
    \caption{\textbf{Meta-Awareness of GRPO vs \metaname{}.} Predicted difficulty and solution length are elicited from both models using the same meta-prediction prompt, and the predictions are parsed from the model outputs. Difficulty is defined by Pass@1 scores, while length refers to the model output token count. Note that jitter is applied to discrete difficulty values to aid density visualization.}
    \label{fig:teaser}
\end{figure}

To this end, we propose a novel RL framework, \longname{} (\metaname{}), which formalizes meta-awareness in reasoning models by rewarding the internal consistency of self-generated signals, thereby eliminating the need for external supervision. Our method introduces a self-predictive trajectory coupled with the primary reasoning path, enhancing the model's meta-awareness of its computational budget, knowledge boundaries, and cognitive strategy. These improved meta-predictions, shown in \Cref{fig:teaser}, drive training efficiency through \textit{predictive gating}, which prunes zero-variance prompts by identifying those that are either trivial or unsolvable, and \textit{early cutoff}, which terminates long rollouts predicted to result in incorrect outcomes. Furthermore, the model could leverage the cognitive strategy to self-generate to provide hints for primary reasoning process.

Building on this foundation, we evaluate the effectiveness of our approach by combining with GRPO and DAPO \citep{yu2025dapo, deepseekmath_grpo}, showing that our method is not dependent on a specific policy gradient algorithm. Remarkably, \metaname{} achieves substantial improvements in mathematical benchmarks with the strongest performance compared under the same compute budget. Finally, predictive gating and early cutoff deliver significant efficiency gains, attaining baseline performance 1.28 times faster than the GRPO training, with a higher accuracy score.

The contributions of this paper can be summarized as follows:
\vspace{-0.05in}
\begin{itemize}[leftmargin=*,itemsep=0mm]
    \item We introduce a predictive reward signal formulated as a parallel verification prompt, which enables the model to self-evaluate meta-awareness by alignment.

    \item We experimentally show that the meta-prediction directly drives performance gain through paired analysis.
     
    \item We propose \effname{}, a post-training strategy with predictive gating and early cutoff, which achieves the strongest performance with minimal training compute.
\end{itemize}

\section{Related Works}
\textbf{Meta-Cognitive Learning}\xspace
Meta-cognition is viewed as a prerequisite for self-improving LLMs \citep{liu2025position}. Existing methods rely on extrinsic mechanisms with fixed action loops, limiting adaptability. Self-improving agents that plan, regulate, and reflect \citep{dong2025meta,didolkar_metacognitive} or refine prompts via past reasoning \citep{qiu2025mela,liu2025ghpo} entangle control with reasoning, often causing interference. In contrast, our approach disentangles the meta and solution path separately for stable training on meta-awareness.

Other works require curated datasets \citep{mera}, or delegate control to external verifiers \citep{ma2025large,he2025good} or multi-agent systems \citep{wan2025rema,yang2025learning,bilal2025meta,khandelwal2025language}, reducing scalability of meta-cognitive training. Training-free heuristics such as confidence-based stopping \citep{yang2025dynamic,qiao2025concise,lu2025prolonged} or correctness checks \citep{ma2025large} offer efficiency but lack genuine language-level meta-cognition. In contrast, our approach does not rely on human-curated reasoning pipelines, external verifiers, PRMs, or specialized datasets targeting meta-cognitive ability, but rather leverages the \textit{self-generated signals to encourage alignment} between the meta-prediction and primary thinking process.

\textbf{Self-Control for Efficient Training}\xspace
Another direction that leverages meta-cognition is to regulate reasoning efficiency by allocating budgets via difficulty assessment \citep{chen2025aware,tu2025learning,shi2025efficient,qu2025optimizing,adactrl,difficultystage,preview_difficult_intervention,han2024token,fang2025thinkless,yang2025think,zhang2025edge,wang2025adaptive,zhang2025adaptthink,shen2025dast}, constraining output length with penalties or fixed limits \citep{lcpo_l1,lmpo,alp,grpo_lead}, and adaptively stopping, continuing, or reflecting for compact reasoning \citep{mera,rlvmr,dai2025s}. While these methods improve inference-time efficiency, they focus on making reasoning shorter or faster at inference time, often at the expense of reasoning performance drop. In contrast, we target \textit{efficiency during the post-training phase}, achieving both efficiency and improved performance during model training rather than the inference.

\section{\metaname{}: \longname{} and \effname{}}
We first provide background on group relative policy optimization (GRPO) (\Cref{subsec:preliminary}). Then we show our method: (i) \metaname{}, which endows the LLM with the capability to perform accurate meta-predictions (\Cref{subsec:self-alignment_rewarding}); and (ii) \effname{}, an efficiency-enhanced version that accelerates \metaname{} through predictive gating and early cutoff. (\Cref{subsec:prediction-based gating and cutoff}).

\subsection{Preliminaries}\label{subsec:preliminary} We present an overview of GRPO, which is a popular RL algorithm for post-training reasoning models.
The old policy model $\pi_{\theta_\text{old}}$ produces a group of $G$ responses given prompt $\mathbf q$ from tasks $P(\mathcal Q)$, creating rollouts $\mathcal O = \{\mathbf{o}_1, \cdots, \mathbf{o}_G\}$. Each response is assigned a reward $\{r_1, \cdots, r_G\}$ based on the rule-based verification of the extracted answer against the ground truth.

The objective of GRPO is formulated as,
\begin{equation}
\resizebox{0.90\linewidth}{!}{$
\begin{split}
    &\mathbb{E}_{\substack{\mathbf{q} \sim P(\mathcal{Q}) \\ \{\mathbf{o}_i\} \sim \pi_{\theta_\text{old}}(\cdot|\mathbf{q})}} \left[ \frac{1}{G}\sum_{i=1}^G \frac{1}{|\mathbf{o}_i|} \sum_{t=1}^{|\mathbf{o}_i|} \left( \mathcal{J}^{\text{clip}}_{i,t}(\theta) - \beta D_{\mathrm{KL}}(\pi_\theta || \pi_{\text{ref}}) \right) \right] \\
    &\text{where } \mathcal{J}^{\text{clip}}_{i,t}(\theta) = \min \Big( \rho_{i,t}(\theta)\hat{A}_{i,t}, \text{clip}\big(\rho_{i,t}(\theta), 1-\epsilon, 1+\epsilon\big)\hat{A}_{i,t} \Big).
\end{split}
$}
\end{equation}

Note that $\rho_{i,t}(\theta) = \frac{\pi_\theta(o_{i,t} |\mathbf{q}, \mathbf{o}_{i,<t})}{\pi_{\theta_{\text{old}}}(o_{i,t}|\mathbf{q}, \mathbf{o}_{i,<t})}$ denotes the importance sampling ratio, and $\pi_{\theta}$ represents the current policy model. $\pi_{\text{ref}}$ is the reference model. $\mathrm{clip}(\cdot)$ restricts the importance sampling ratio between $[1-\epsilon, 1+\epsilon]$. The advantage is calculated as $\hat{A}_{i,t} = \frac{r_i - \mathrm{mean}(\{r_i\}_{i=1}^G)}{\mathrm{std}(\{r_i\}_{i=1}^G)}$. Following the practice of recent GRPO variants \citep{drgrpo, grpo_lead, zheng2025group, yu2025dapo}, we set $\beta=0$ to remove the KL divergence term.

\subsection{\metaname{}: Designing \longname{}}\label{subsec:self-alignment_rewarding}
\paragraph{Overall Pipeline of \metaname{}} Building on the GRPO-based framework, the policy model is prompted with two distinct inputs: \textit{solution} prompt $\mathbf{q}_\text{sol}$ and \textit{meta} prompt $\mathbf{q}_\text{meta}$. The solution and meta rollouts are executed simultaneously, but the rewarding pipelines differ. Solution rollouts are verified against static ground truth using rule-based verification, while meta rollouts are verified against empirical statistics derived from the solution rollouts as dynamic ground truth.

The \textit{solution} prompt $\mathbf{q}_\text{sol}$ instructs the model to solve the problem via chain-of-thought, generating a group of $G$ solution rollouts as detailed in \Cref{subsec:preliminary}. For verification on meta-prediction, the average Pass@1 score over $G$ rollouts ($p$), the range of output token length from correct rollouts ($[l_\text{min}, l_\text{max}]$) is extracted, and the entire responses ($\mathcal O$) are saved for predicted notion verification.

Simultaneously, the \textit{meta} prompt $\mathbf{q}_\text{meta}$, instructs the model to predict the expected difficulty as Pass@1 score ($\hat p$), the expected length of correct response ($\hat l$), and a set of problem-solving notions ($\hat{\mathcal G}_\text{notion}$). We generate $M$ independent meta-rollouts,\footnote{The full meta-prediction prompt template is deferred to \Cref{sec:appendix}.} and reward each by how accurately it predicts the output length, problem difficulty, and used notions from the solution rollouts. The meta rewards, $\{r^{\text{meta}}_1, \dots, r^{\text{meta}}_M\}$, are then normalized within the group of $M$ rollouts to compute advantages. The reward computation for each meta component is detailed below. For reproducibility, we provide the complete code snippet in \Cref{app:reward}.

\paragraph{Difficulty Reward.} The difficulty alignment reward measures the proximity between the predicted pass-rate $\hat p$ and the actual pass-rate $p$. This is the proportion of correct answers among $G$ rollouts for question $q$. This allows the model to learn how hard the given question is for the current knowledge boundary of the model.
 
We compute the accuracy score as an exponential decay function of the normalized prediction error, given by \[r_{\text{difficulty}} = 0.01^{|p - \hat{p}|}.\] 

A deviation of a single unit in difficulty prediction approximately halves the reward with the base number 0.01, in order to strongly penalize higher errors in prediction.

\paragraph{Length Reward.} The length alignment reward checks if the predicted length falls within the range of correct responses. Formally, we assign the reward if the predicted length $\hat l$ falls in-between the min-max range of correct responses as
\[r_{\mathrm{length}} = \mathbbm{1} \big[ l_\text{min}  \leq \hat l \leq l_\text{max} \big].\] 
If no correct solution exists, then we set the reward as 0.

\paragraph{Notion Reward.} The notion reward evaluates whether the predicted problem-solving notion emerges more frequently in correct rollouts than in incorrect ones. More formally, for a single notion $n \in \hat{\mathcal{G}}_{\text{notion}}$, we count the number of correct responses containing $n$ (denoted as $c_{\text{corr, n}}$) and the number of incorrect responses containing $n$ (denoted as $c_{\text{wrong, n}}$).

Then, the notion reward is defined as
\[r_\text{notion} = \mathbb E_{n \sim \hat {\mathcal G}_\text{notion}}\big[  \mathbbm{1}[c_\text{corr, n} > c_\text{wrong, n}]  \big] .\]
Notions present in the problem statement are excluded to prevent reward hacking, and lemma-based matching is used for counting.

Then, the meta reward is defined as the average of three componenets,
\begin{equation}\label{eq:meta_reward}
r^{\mathrm{meta}} = \frac{r^{\mathrm{length}} + r^{\mathrm{difficulty}} + r^{\mathrm{notion}}}{3}.
\end{equation}

\begin{figure}[t]
    \centering
    \includegraphics[width=\linewidth]{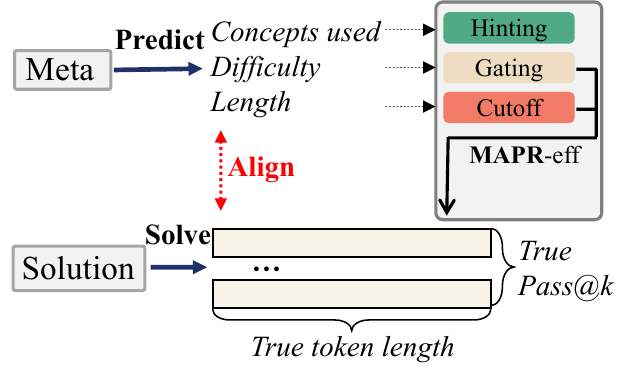}
     \caption{\textbf{Overall Framework of \metaname and \effname} \metaname{} predicts and solves in parallel from given \textit{meta} and \textit{solution} prompts. The predicted values are verified against true pass@k, token length, and used concepts extracted from the solution rollouts. The efficient version, \effname{}, applies predictive gating and length cutoff for efficient training.}
     \label{fig:framework}
\end{figure}

\begin{table*}[t!]
    \small
    \setlength{\extrarowheight}{3pt}
    \setlength{\tabcolsep}{15pt}
    \centering
        \caption{\textbf{Performance of GRPO and \metaname{} for Math benchmarks.} Pass@1 and Pass@8 scores are reported with standard deviations over 32 samplings. The overall performance of our method \metaname{} surpasses baseline GRPO method by large margin.}
        \label{tab: comparison on six benchmarks on qwen3-14b and qwen3-8b}
    \resizebox{0.88\linewidth}{!}{
    \begin{tabular}{l|cc|cc}
    \toprule
        \multirow{2}{*}{\textbf{Benchmark}} & \multicolumn{2}{c|}{\textbf{GRPO}} & \multicolumn{2}{c}{\textbf{GRPO w/ \metaname{}}} \\
            & Pass@1 & Pass@8  &  Pass@1 & Pass@8 \\
        \hline 
        \rowhighlight
        \multicolumn{5}{c}{\textbf{Qwen3-4B Base Model}} \\
        \hline
        AIME'24       & 17.50\scriptsize{$\pm$4.00} & 33.60\scriptsize{$\pm$5.96} & \textbf{26.15\scriptsize{$\pm$3.32}}  \color{cadmiumgreen}{(+ 49.43\%)} & \textbf{48.82\scriptsize{$\pm$5.32}} \color{cadmiumgreen}{(+ 45.30\%)}\\
        AIME'25       & 11.77\scriptsize{$\pm$4.56} & 25.56\scriptsize{$\pm$4.40} & \textbf{21.56\scriptsize{$\pm$4.40}} \color{cadmiumgreen}{(+ 83.18\%)} & \textbf{37.17\scriptsize{$\pm$3.63}}  \color{cadmiumgreen}{(+ 45.42\%)} \\
        AMC23        & 59.30\scriptsize{$\pm$6.40} & 84.93\scriptsize{$\pm$3.90} & \textbf{70.16\scriptsize{$\pm$4.78}} \color{cadmiumgreen}{(+ 18.11\%)} & \textbf{93.18\scriptsize{$\pm$1.90}} \color{cadmiumgreen}{(+ 9.71\%)}\\
        MATH500      & 79.61\scriptsize{$\pm$0.91} & 90.12\scriptsize{$\pm$0.59} & \textbf{84.52\scriptsize{$\pm$0.74}} \color{cadmiumgreen}{(+ 6.17\%)} & \textbf{93.74\scriptsize{$\pm$0.42}} \color{cadmiumgreen}{(+ 4.02\%)}\\
        Minerva  & \textbf{42.27\scriptsize{$\pm$1.53}} & 59.70\scriptsize{$\pm$0.91} & 41.12{\scriptsize{$\pm$2.00}}  \color{cornellred}{(- 3.18\%)}& \textbf{63.78\scriptsize{$\pm$1.35}} \color{cadmiumgreen}{(+ 6.83\%)}\\
        Olympiad & 44.47\scriptsize{$\pm$1.04} & 61.99\scriptsize{$\pm$0.61} & \textbf{53.38\scriptsize{$\pm$0.96}}\color{cadmiumgreen}{ (+ 20.04\%)} & \textbf{69.74\scriptsize{$\pm$0.69}}\color{cadmiumgreen}{ (+ 12.50\%)} \\
        \hline
        \textbf{Average} & 42.49{\scriptsize{$\pm$3.07}} & 59.31{\scriptsize{$\pm$2.73}} & \textbf{49.48{\scriptsize{$\pm$2.70}}} \color{cadmiumgreen}{(+ 13.04\%)} & \textbf{\textbf{67.73}\scriptsize{$\pm$2.22}}{\color{cadmiumgreen}{(+ 14.20\%)}} \\
        \hline 
        \rowhighlight
        \multicolumn{5}{c}{\textbf{Qwen3-8B Base Model}} \\
        \hline
        AIME'24 & 28.54\scriptsize{$\pm$4.12} & 53.96\scriptsize{$\pm$4.07}  & \textbf{34.17\scriptsize{$\pm$5.54}} {\color{cadmiumgreen}{(+ 19.72\%)}} & \textbf{63.80\scriptsize{$\pm$4.98}} {\color{cadmiumgreen}{(+ 18.24\%)}}\\
        AIME'25 & 22.19\scriptsize{$\pm$3.63} & 38.74\scriptsize{$\pm$4.05} & \textbf{28.44\scriptsize{$\pm$5.41}} {\color{cadmiumgreen}{(+ 28.17\%)}} & \textbf{45.96\scriptsize{$\pm$4.41}} {\color{cadmiumgreen}{(+ 18.64\%)}} \\
        AMC23  & 73.67\scriptsize{$\pm$5.60} & 92.77\scriptsize{$\pm$2.43} & \textbf{79.53\scriptsize{$\pm$4.26}} {\color{cadmiumgreen}{(+ 7.95\%)}} & \textbf{94.39\scriptsize{$\pm$1.80}} {\color{cadmiumgreen}{(+ 1.75\%)}} \\
        MATH500 & 85.75\scriptsize{$\pm$0.66} & 94.31\scriptsize{$\pm$0.49} & \textbf{88.05\scriptsize{$\pm$0.82}} {\color{cadmiumgreen}{(+ 2.68\%)}} & \textbf{95.35\scriptsize{$\pm$0.49}} {\color{cadmiumgreen}{(+ 1.1\%)}} \\
        Minerva & 43.21\scriptsize{$\pm$2.12} & 64.00\scriptsize{$\pm$1.14} & \textbf{47.21\scriptsize{$\pm$1.74}} {\color{cadmiumgreen}{(+ 9.26\%)}} & \textbf{68.21\scriptsize{$\pm$1.23}} {\color{cadmiumgreen}{(+ 6.58\%)}} \\
        Olympiad & 54.03\scriptsize{$\pm$1.22} & 70.04\scriptsize{$\pm$0.70} & \textbf{56.86\scriptsize{$\pm$0.85}} {\color{cadmiumgreen}{(+ 5.24\%)}} & \textbf{71.87\scriptsize{$\pm$0.51}} {\color{cadmiumgreen}{(+ 2.61\%)}} \\
        \hline
        \textbf{Average} & 51.23\scriptsize{$\pm$2.89} & 68.97\scriptsize{$\pm$2.15} & \textbf{55.71\scriptsize{$\pm$3.10}} \color{cadmiumgreen}{(+ 8.74\%)} & \textbf{73.26\scriptsize{$\pm$2.24}} \color{cadmiumgreen}{(+ 6.22\%)} \\
        \hline 
        \rowhighlight
        \multicolumn{5}{c}{\textbf{Qwen3-14B Base Model}} \\
        \hline
        AIME'24 & 38.54\scriptsize{$\pm$4.30} & 58.55\scriptsize{$\pm$4.07} &  \textbf{44.27\scriptsize{$\pm$5.64}} {\color{cadmiumgreen}{(+ 14.87\%)}} & \textbf{68.30\scriptsize{$\pm$3.57}} {\color{cadmiumgreen}{(+ 16.65\%)}} \\
        
        AIME'25 & 27.92\scriptsize{$\pm$4.69} & 45.56\scriptsize{$\pm$3.87} & \textbf{31.25\scriptsize{$\pm$5.12}} {\color{cadmiumgreen}{(+ 11.93\%)}} & \textbf{53.57\scriptsize{$\pm$5.51}} {\color{cadmiumgreen}{(+ 17.58\%)}} \\
        
        AMC23 & 81.56\scriptsize{$\pm$4.98} & \textbf{96.20\scriptsize{$\pm$1.66}} & \textbf{86.02\scriptsize{$\pm$4.16}} {\color{cadmiumgreen}{(+ 5.47\%)}} & {95.12\scriptsize{$\pm$1.48}} {\color{cornellred}{(- 1.12\%)}} \\
        
        MATH500 & 88.73\scriptsize{$\pm$1.03} & 96.02\scriptsize{$\pm$0.36} & \textbf{89.93\scriptsize{$\pm$0.88}} {\color{cadmiumgreen}{(+ 1.35\%)}} & \textbf{96.39\scriptsize{$\pm$0.34}} {\color{cadmiumgreen}{(+ 0.38\%)}} \\
        
        Minerva & 45.03\scriptsize{$\pm$1.73} & 66.42\scriptsize{$\pm$1.06} & \textbf{50.36\scriptsize{$\pm$1.53}} {\color{cadmiumgreen}{(+ 11.84\%)}} & \textbf{69.20\scriptsize{$\pm$0.96}} {\color{cadmiumgreen}{(+ 4.19\%)}} \\
        
        OlympiadMath & 59.04\scriptsize{$\pm$0.90} & 73.03\scriptsize{$\pm$0.58} & \textbf{61.59\scriptsize{$\pm$0.89}} {\color{cadmiumgreen}{(+ 4.32\%)}} & \textbf{74.37\scriptsize{$\pm$0.65}} {\color{cadmiumgreen}{(+ 1.83\%)}} \\
        \hline 
        \textbf{Average} & 56.80\scriptsize{$\pm$2.94} & 72.63\scriptsize{$\pm$1.93} &\textbf{60.57\scriptsize{$\pm$3.04}} \color{cadmiumgreen}{(+ 6.63\%)} & \textbf{76.15\scriptsize{$\pm$2.09}} \color{cadmiumgreen}{(+ 4.85\%)} \\
    \bottomrule
    \end{tabular}
    }
\end{table*}

\subsection{\effname{}: Meta-based Active Control for Efficient Post-Training}\label{subsec:prediction-based gating and cutoff}
\effname{} is a variant of \metaname{} that can further boost training efficiency by leveraging the length and difficulty predictions. 

\paragraph{Overall Pipeline of \effname{}}
To encourage meta-awareness before accelerating the training phase, we first perform self-alignment based policy updates for the early $k$ steps of update with self-predictive alignment reward, until the policy model shows stable meta-prediction alignment with the true solution rollouts. After $k$-th step, we alter into non-parallel pipeline that executes meta-predictions first, for {predictive gating}, followed by solution rollouts, applying {early length cutoff}. We may also utilize the predicted notions to provide additional hint for the model in solving the questions.

\textbf{Predictive gating} acts as a pre-computation filter for tasks that are deemed either trivial or impossible to solve. For a given question $q$, gating engages only when the standard deviation across $M$ predicted pass-rates falls below $\sigma_\text{pg}$ and the average prediction is 0 or 1. Distinct from methods like DAPO, which prune \textit{after} expensive solution rollouts, our approach conserves computation by gating \textit{before} the rollout phase. Since our primary objective is training efficiency, we employ static online gating; dynamically re-evaluating previously gated tasks would require periodic, costly meta-predictions on excluded data. 

\textbf{Length cutoff} restricts generation to the predicted length, scaled by a margin $l_\text{LC}$. As the \metaname{} length reward incentivizes accurate prediction for correct rollouts, exceeding this threshold is highly unlikely to yield a correct answer, despite the cost of generating additional tokens. Therefore, \effname{} utilizes the length prediction as a hard threshold to terminate rollouts once the limit is reached. 

Additionally, \textbf{notion feedin} is implemented by appending the hint ``The problem could be solved using the following math notions'' to the problem statement, providing auxiliary guidance during the solution rollout phase.

\begin{figure*}[t]
    \centering
    \begin{subfigure}[t]{0.54\textwidth}
        \centering
        \includegraphics[width=\linewidth]{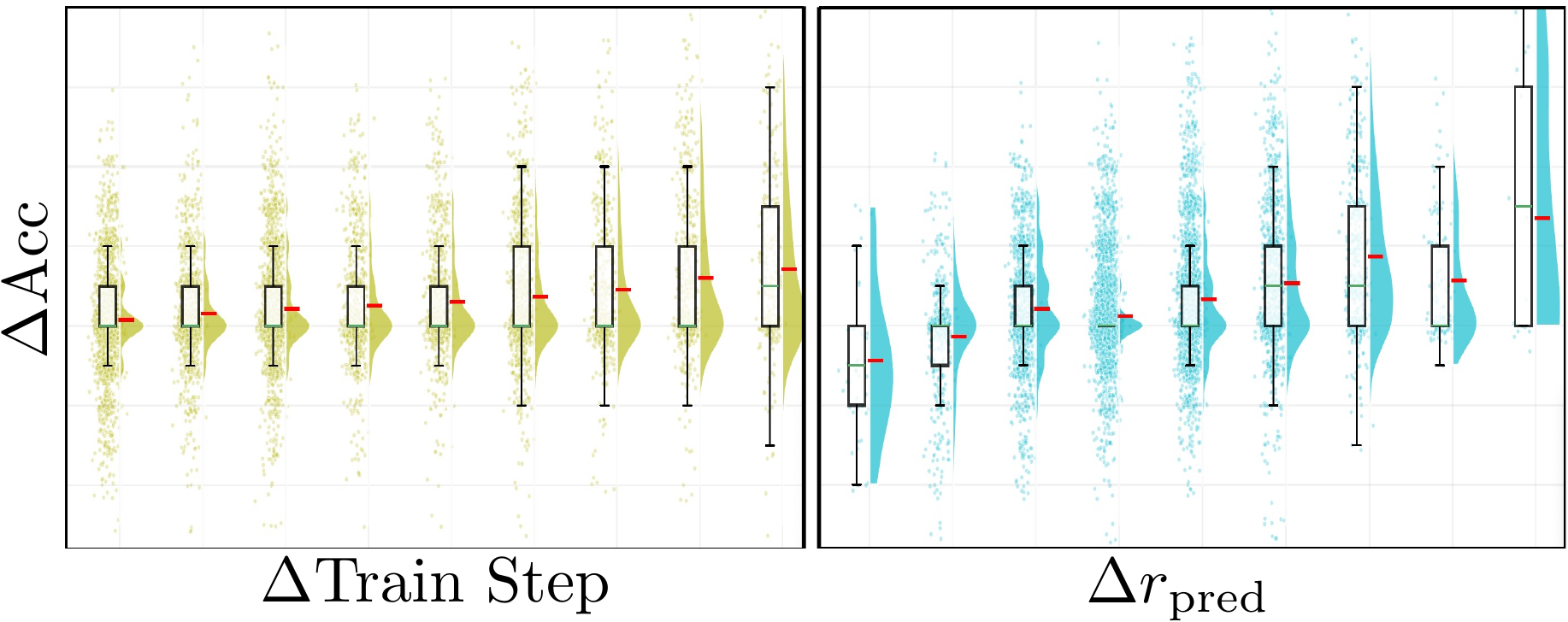}
        \caption{Sensitivity Analysis}
        \label{fig:sensitivity}
    \end{subfigure}
    \hfill 
    \begin{subfigure}[t]{0.44\textwidth}
        \centering
        \includegraphics[width=\linewidth, trim={0 1.1cm 0 0}, clip]{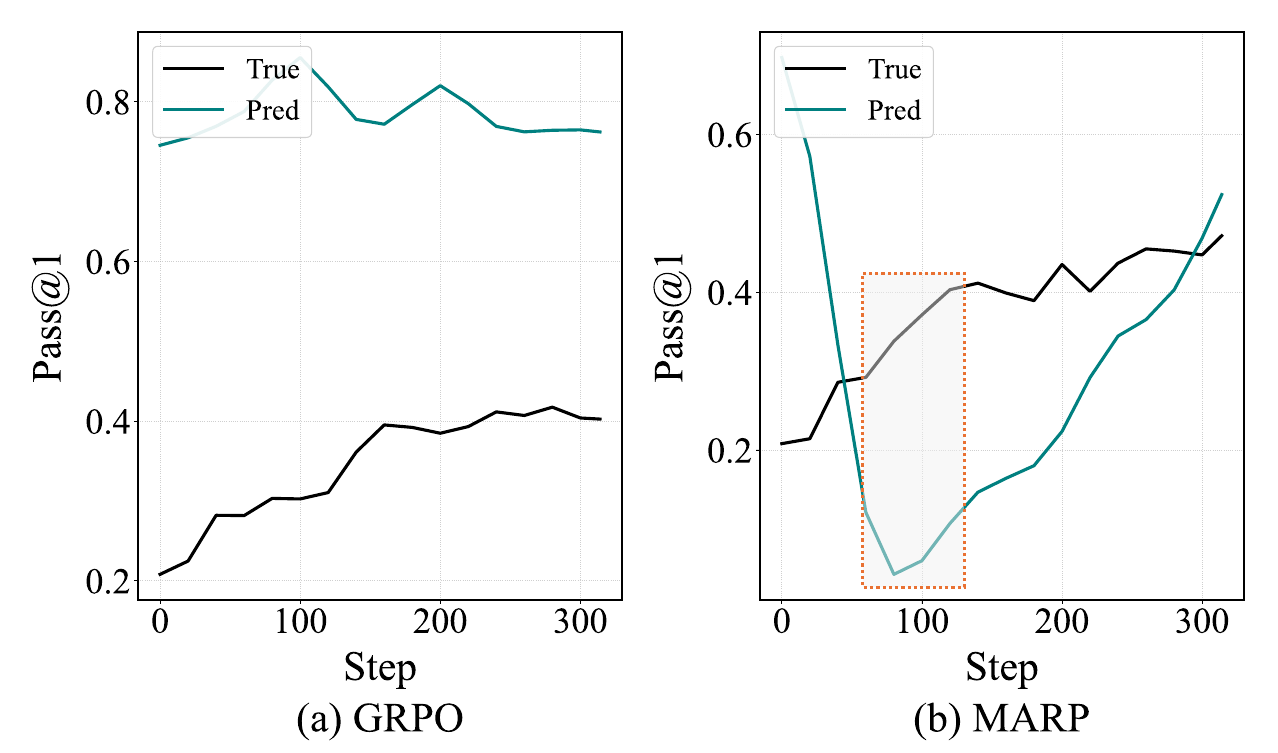}
\makebox[0.48\linewidth][c]{\textbf{GRPO}} \makebox[0.48\linewidth][c]{\textbf{MAPR}}
        \caption{Alignment of Difficulty}
        \label{fig:meta-acc}
    \end{subfigure}
    
    \vspace{-5pt} 
    
    \caption{\textbf{Impact of Meta-Awareness on Training Dynamics.} 
    (a) We observe a significantly steeper gradient for meta-awareness ($r_\text{pred}$) compared to training steps, suggesting that increased meta-awareness drives performance more effectively than training duration alone. 
    (b) The \metaname{} Pass@1 surge (steps 80-120) coincides precisely with the \textit{drop-then-align} phase in difficulty prediction (orange), implying that predictive calibration correlates strongly with performance increase.}
    \label{fig:combined_analysis}
\end{figure*}

\section{Experiments}
In this section, we provide the details of training and evaluation configuration in \cref{subsec:eval detail}. Then we demonstrate the performance gain and efficiency driven by \metaname{} and \effname{} in \cref{subsec:apo analysis}. In addition, we systematically analyze the components of our method through ablation studies in \cref{subsec:ablation}.

\subsection{Train and Evaluation Details}\label{subsec:eval detail}

\textbf{Training Details.}\xspace We use VeRL with the DeepScaleR~\citep{deepscaler2025} dataset, batch size 128, learning rate 1e-6, $10\%$ weight decay, maximum response length 8K, and GRPO without KL term. Training runs for one epoch (314 steps) using AdamW~\citep{adamw} with 20 warm-up steps, gradient clipping at 1.0, and clipping range for GRPO between $[\epsilon_\mathrm{low}=0.2,\epsilon_\mathrm{high}=0.28]$. The rollouts use temperature 1.0 and top-p value of 1.0. The number of rollouts is 16 for the response generation, and 8 for meta prediction.

\textbf{Evaluation Configuration.}\xspace
We use the provided math scoring function in VeRL to measure the accuracy of the predicted answer and ground truth answer, sampling 32 responses, with 16k maximum response length and temperature set at 0.6.

We evaluate the performance of our method using six widely used mathematical reasoning benchmarks, AIME24, AIME25, AMC23, MATH500~\citep{math500}, Minerva, and OlympiadBench~\citep{olympiadbench}. Experiments are conducted on Qwen3 8B base model unless otherwise stated.

\subsection{Analysis on \metaname{} and \effname{}}\label{subsec:apo analysis}
\paragraph{\metaname{} Excels in Math Benchmark} \metaname{} excels the baseline in six math benchmarks - AIME24, AIME25, AMC23, MATH500, Minerva, and OlympiadBench (\Cref{tab: comparison on six benchmarks on qwen3-14b and qwen3-8b}). Across all mathematical datasets, our method \metaname{} shows great improvement over the baseline GRPO performance, showing an average of 13.04\% of improvement in Qwen3-4B model, 8.74\% in Qwen3-8B model, and 6.63\% in Qwen3-14B model. Among the six benchmarks, \metaname{} gains maximum performance on intermediate to hard level (AIME, AMC, Olympiad, Minerva), while the performance boost for MATH500 shows performance saturation especially for large scaled model of 14B. We also demonstrate the superior ability of \metaname{} on out-of-domain benchmarks, ranging from logical, scientific, to coding domains in \Cref{tab:generalization}.

\paragraph{Meta-Awareness Directly Enhances Performance}

In \Cref{fig:sensitivity}, we assess whether performance gains arise from improvements in the reward metric $r_\text{pred}$ or from extended training. We conduct a paired analysis comparing marginal accuracy gains ($\Delta \text{Acc}$) with both training steps and meta-awareness measured by $r_\text{pred}$. Checkpoints are sampled every 20 steps across six mathematical benchmarks, and for each question we pair model states $(t, q)$ from different steps to control for input variation.

We plot step differences versus accuracy gains using a jittered distribution. To isolate the effect of meta-awareness, we compute $\Delta r_\text{pred} = |r_\text{pred}^t - r_\text{pred}^q|$ and plot binned values against $\Delta \text{Acc}$. Accuracy improves more steeply with respect to $r_\text{pred}$ than with training steps, indicating that performance is more sensitive to meta-cognitive calibration than to additional training compute. 

\paragraph{Meta-prediction Dynamics During \metaname{} Training}
As shown in \Cref{fig:meta-acc}, a critical divergence in training dynamics appears when analyzing the model's self-prediction of problem difficulty ($\hat p$) versus the true difficulty ($p$). GRPO exhibits consistent overconfidence, predicting a Pass@1 value exceeding 0.8, despite its true score remaining significantly lower. 

In contrast, while \metaname{} also begins with initial overconfidence, the meta-awareness objective drives a corrective \textit{drop-then-align} behavior in \Cref{fig:meta-acc}. The predicted difficulty drops sharply until step 80, recalibrating to match the true Pass@1. Crucially, this coincides with the rapid ascent in the true Pass@1 score, suggesting that accurate self-assessment correlates with the performance gains observed in \metaname{}. Similar tendency is also observed in length prediction ($l$ vs $\hat l$), which is shown in \Cref{app:len_tendency}.

\paragraph{\effname{} Achieves the Strongest Performance with Minimal Training Compute}

\begin{figure}[t]
    \includegraphics[width=\linewidth]{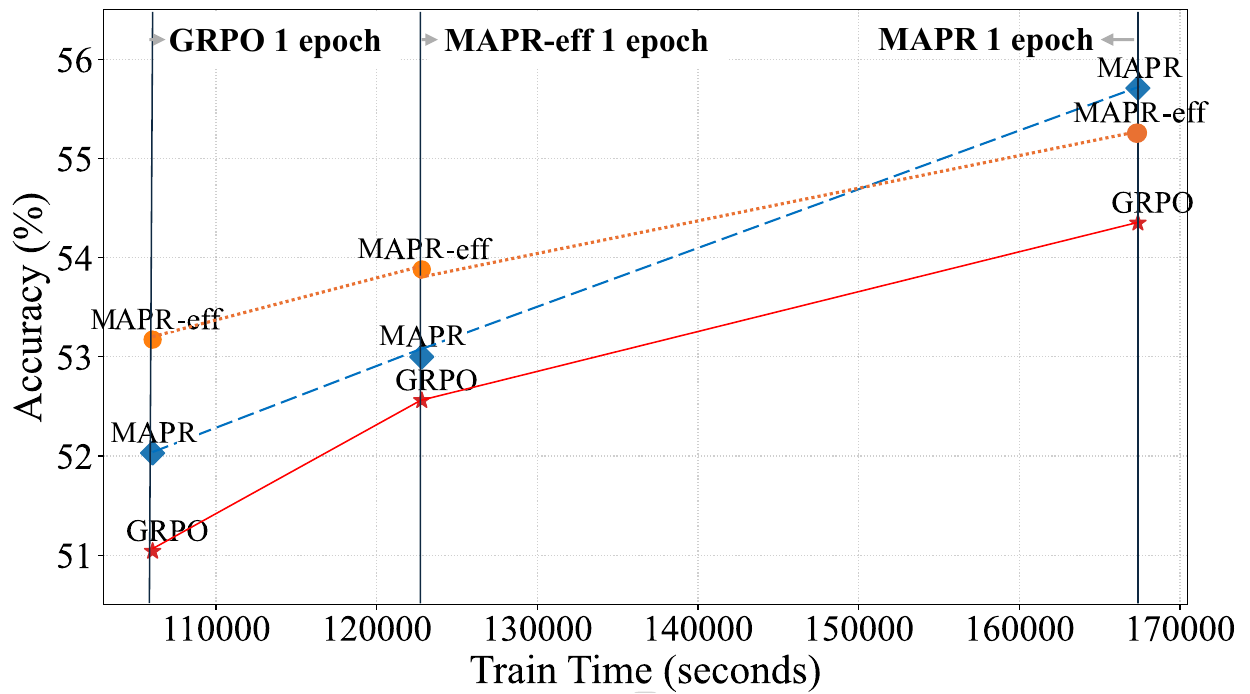}
    \caption{\textbf{Accuracy vs Wall Clock Time.} Average performance across six math benchmarks. Gray vertical lines indicate epoch milestones. Both \effname{} and \metaname{} achieve Pareto-superiority over the GRPO baseline, showing higher accuracy for the same compute expenditure.}
    \label{fig:pareto}
\end{figure}

Under the same compute budget (wall clock time), we demonstrate that \metaname{} and \effname{} surpass the performance of GRPO. In \cref{fig:pareto}, we demonstrate the average accuracy over six mathematical benchmarks at three different timestamps, which are 1 epoch (314 steps) duration of three model variants GRPO, \effname{}, and \metaname{}, emphasized as gray vertical lines on the plot. Under the same train compute at three different timestamps, \effname{} and \metaname{} consistently outperform the baseline method GRPO by a large margin. This proves the efficacy of our method in achieving a large performance gain even under same amount of training compute time.

\begin{figure}[t]

    \centering
    \begin{minipage}[c]{0.3\linewidth}
        \centering
        \label{tab:pg_metrics}
        \resizebox{\linewidth}{!}{%
            \begin{tabular}{lc}
                \toprule
                \textbf{Metric} & \textbf{Value} \\
                \midrule
                Precision & 0.9417 \\
                Recall    & 0.8739 \\
                F1 Score  & 0.9065 \\
                \bottomrule
            \end{tabular}%
        }
    \end{minipage}
    \hfill
    \begin{minipage}[c]{0.64\linewidth}
        \centering
        \includegraphics[width=\linewidth,trim={0 1.2cm 1.3cm 1cm}, clip]{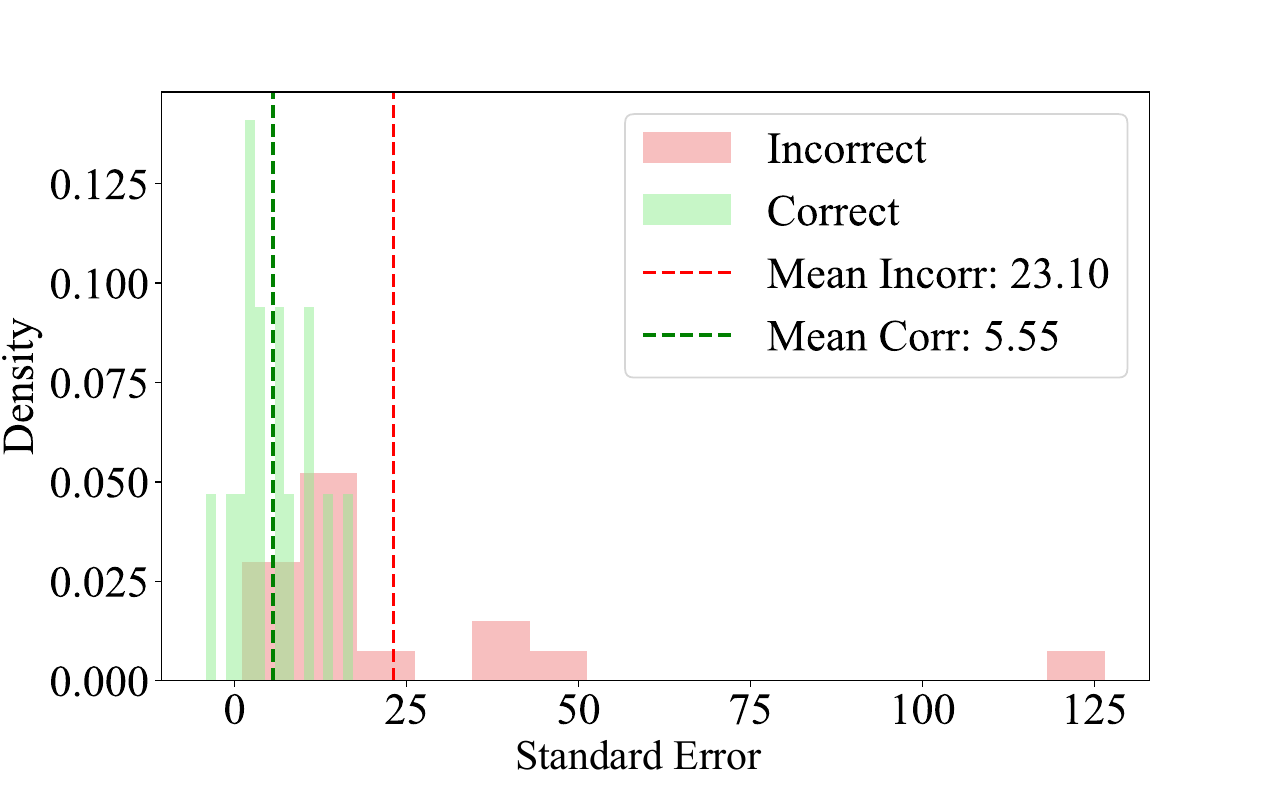}
        \label{fig:lc_correct}
    \end{minipage}
     \makebox[0.34\linewidth][c]{(a) Predictive Gating} \makebox[0.65\linewidth][c]{(b) Length Cutoff}
     \caption{(a) Accuracy of Predictive Gating (b) Standard Error of Length Cutoff from \effname{}.}
     \label{fig:lc-pg-reliability}
\end{figure}

\paragraph{Prediction Performance of PG and LC}
To evaluate the reliability of predictive gating and length cutoff in \effname{}, we compare gating and cutoff decisions against the ground-truth. Using unseen part of DeepScaleR train dataset, \Cref{fig:lc-pg-reliability}\textcolor{cornellred}{a} reports the performance of predictive gating in terms of precision, recall, and F1 score against true zero-variance questions. These metrics evaluate whether the predicted difficulty value of 0 or 1 with a standard deviation below $\sigma_\text{pg}$ matches the true zero variance. Moreover, \Cref{fig:lc-pg-reliability}\textcolor{cornellred}{b} shows the standardized error value of the length cutoff decision. The standard error is calculated as $\left(\mathbb E_{i\sim \mathcal O_\text{sol}}(l_i) - \mathbb E_{i\sim \mathcal O_\text{meta}}(\hat l_i)\right) / \sigma(\hat l)$, which quantifies the deviation of the true length from the prediction, normalized by meta-prediction uncertainty. The distribution shows the standard error from correct rollouts (green) and incorrect rollouts (red). While the distribution for correct rollouts are centered around zero error (0), incorrect rollouts are distributed in larger values. This demonstrates that the length predictions are highly accurate, and the cutoff strategy effectively prevents the model from generating futile extra tokens that would lead to wrong answers. The distribution of standard error values toward the positive range indicates that the length cutoff mechanism serves as a highly effective means of conserving tokens.

\begin{figure*}[t]
    \centering
    \begin{subfigure}[t]{0.28\linewidth}
        \centering
        \includegraphics[width=\linewidth]{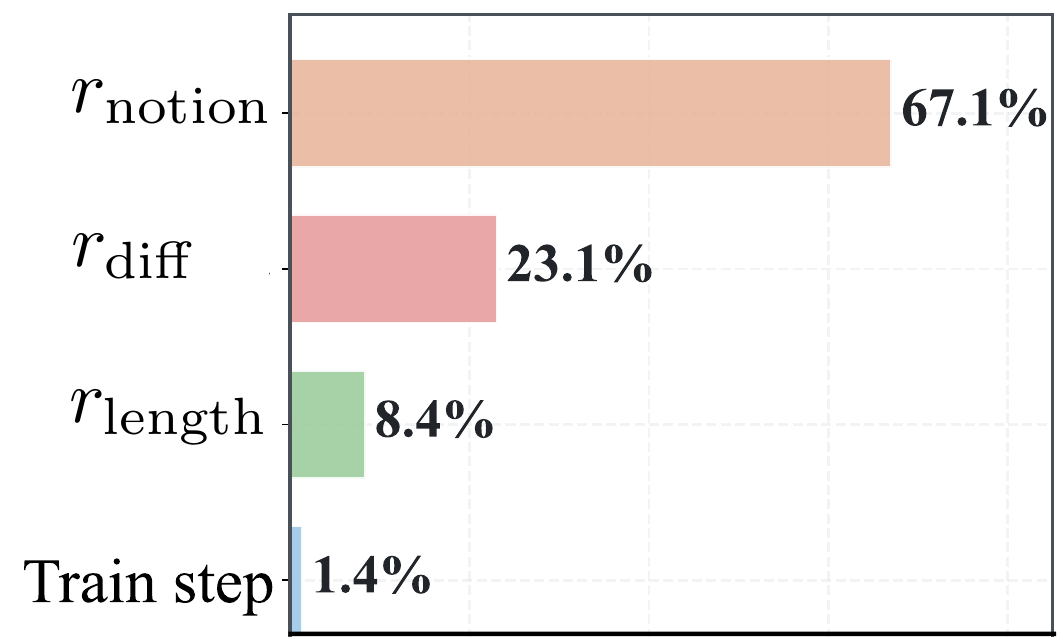}
        \caption{Shapley $R^2$ Analysis on component-wise contributions.}
        \label{fig:Shapley}
    \end{subfigure}
    \hfill
    \begin{subfigure}[t]{0.26\linewidth}
        \centering
        \includegraphics[width=\linewidth]{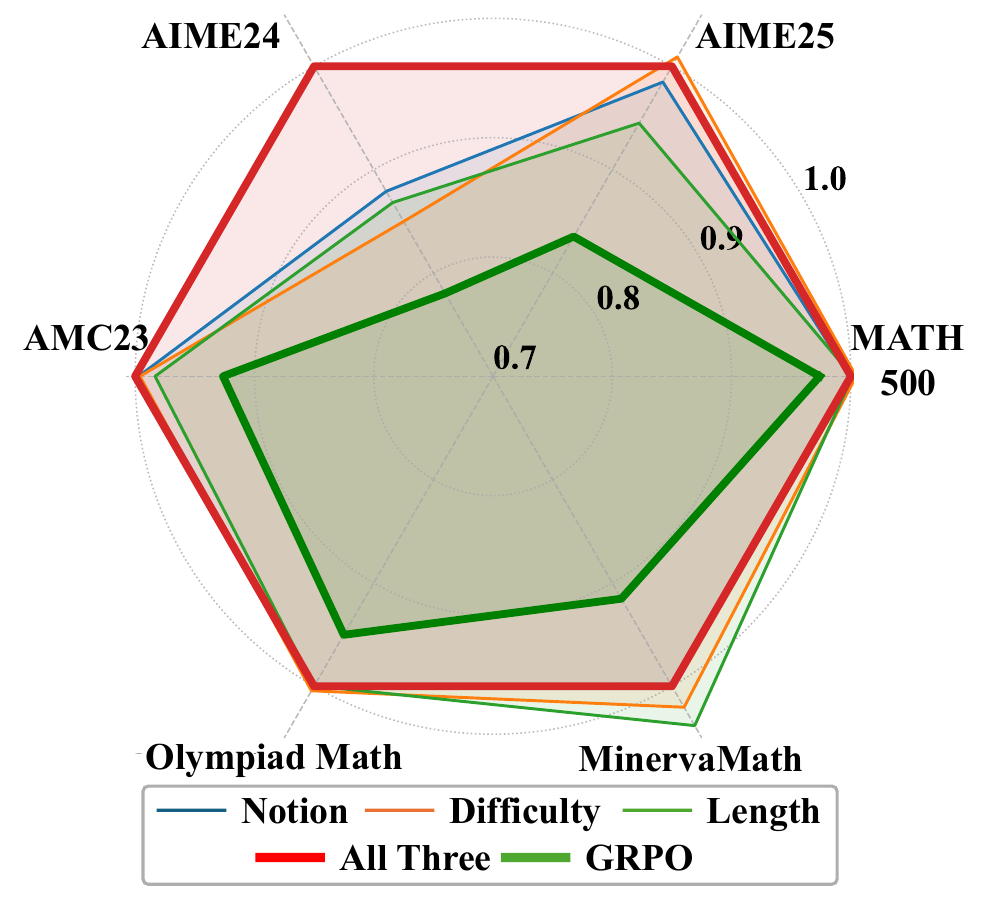}
        \caption{Ablation of meta components (Maximum set to the score of `All three' components for visualization purpose).}
        \label{fig:radar}
    \end{subfigure}
    \hfill
    \begin{subfigure}[t]{0.44\linewidth}
        \centering
        \includegraphics[width=\linewidth]{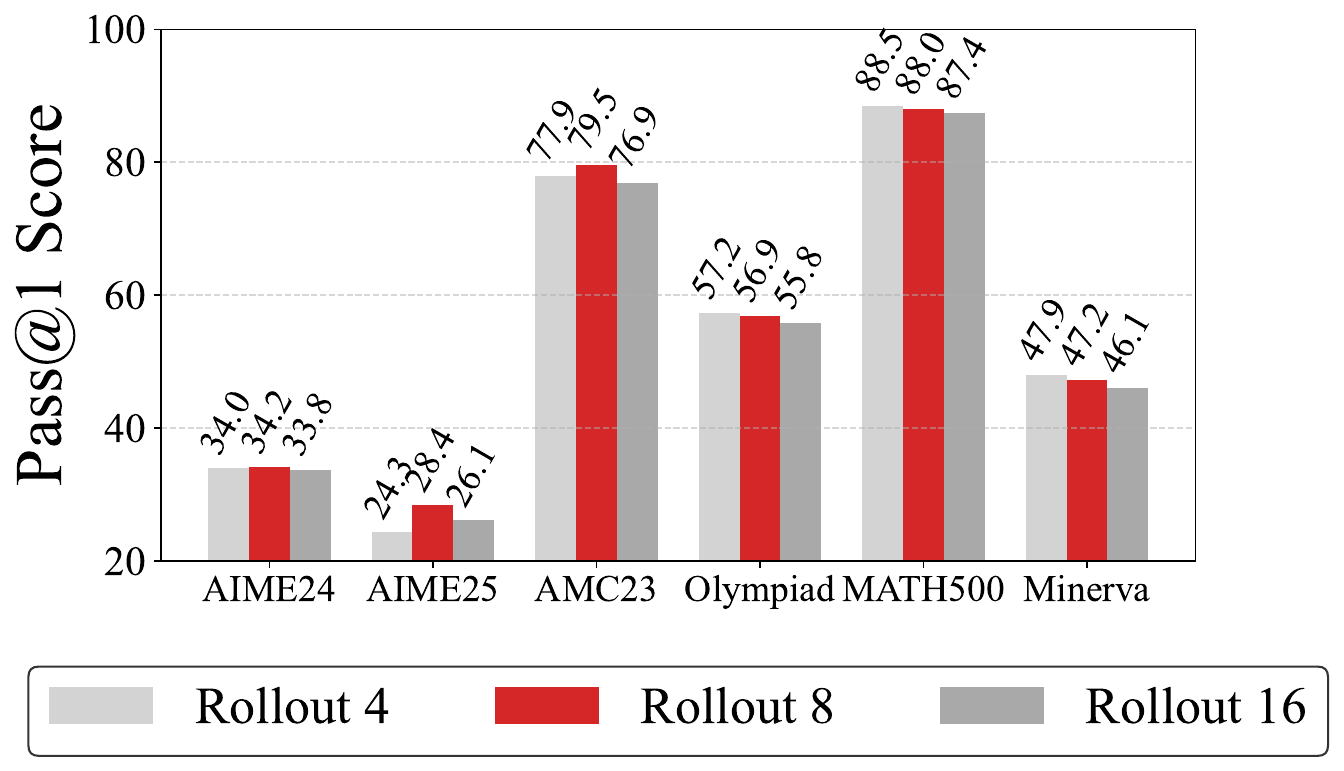}
        \caption{Ablation on number of meta-prediction rollouts.}
        \label{fig:n_rollout}
    \end{subfigure}
    \caption{\textbf{Component Analysis and Ablation Studies} The contribution of our meta-aware predictive reward components analyzed through Shapley values (left), meta type ablation performance (center), and the performance over different numbers of meta rollouts (right).}
    \label{fig:combined_ablations}
\end{figure*}

\begin{figure*}[htp]
\label{fig:ablation_combined}
    \centering
    \begin{subfigure}[b]{0.3\textwidth}
        \centering
        \includegraphics[width=\textwidth, width=\textwidth, trim=0.8cm 0.8cm 0.8cm 0cm, clip]{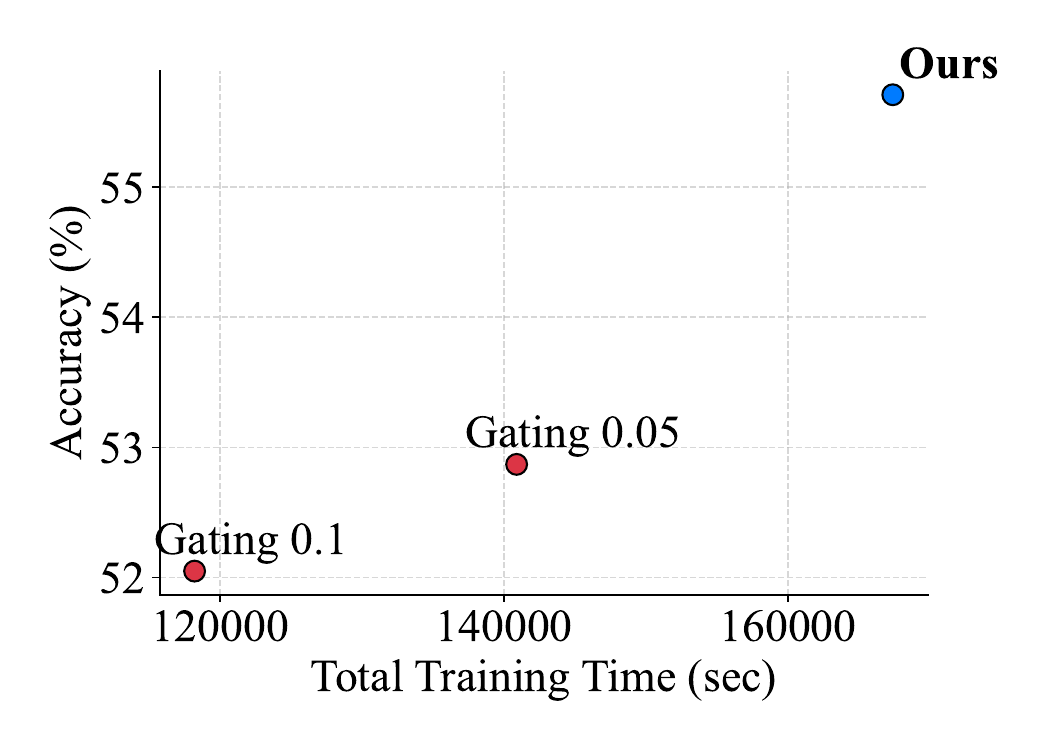}
        \caption{Gating $\sigma_\text{PG}$ Ablation.}
        \label{fig:pg}
    \end{subfigure}
    \hfill 
    \begin{subfigure}[b]{0.3\textwidth}
        \centering
        \includegraphics[width=\textwidth, trim=0.8cm 0.8cm 0.8cm 0cm, clip]{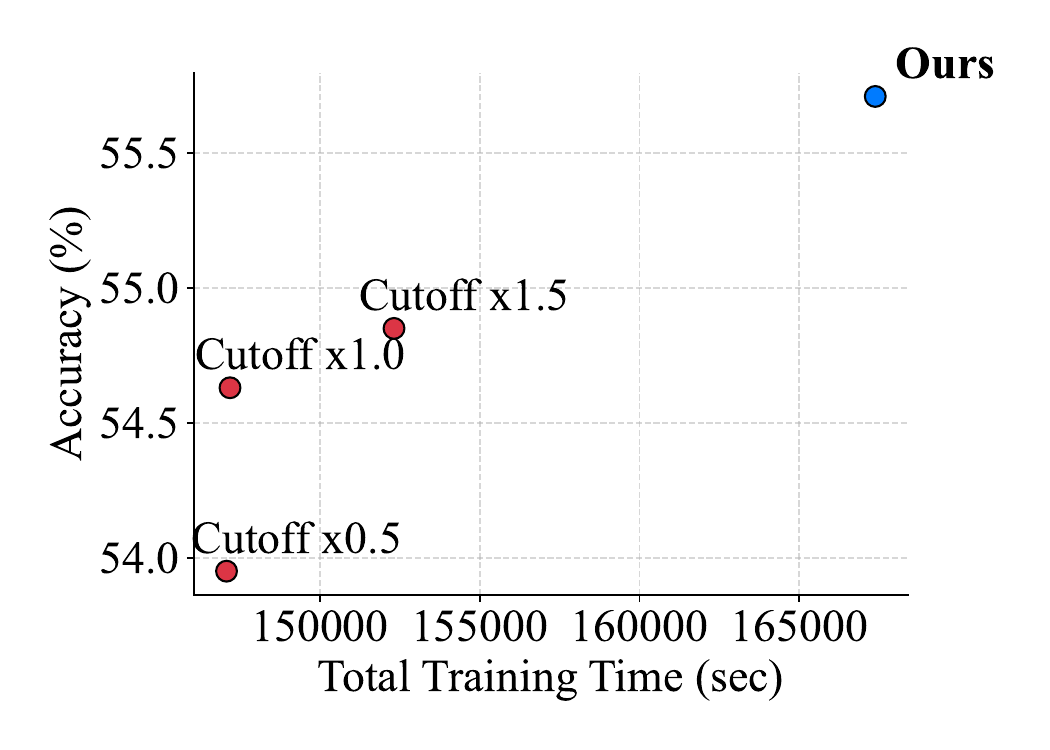}
        \caption{Cutoff $l_\text{LC}$ Ablation.}
        \label{fig:lc}
    \end{subfigure}
    \hfill
    \begin{subfigure}[b]{0.38\textwidth}
        \centering
        \includegraphics[width=\textwidth, trim=0.3cm 0.6cm 0.3cm 0cm, clip]{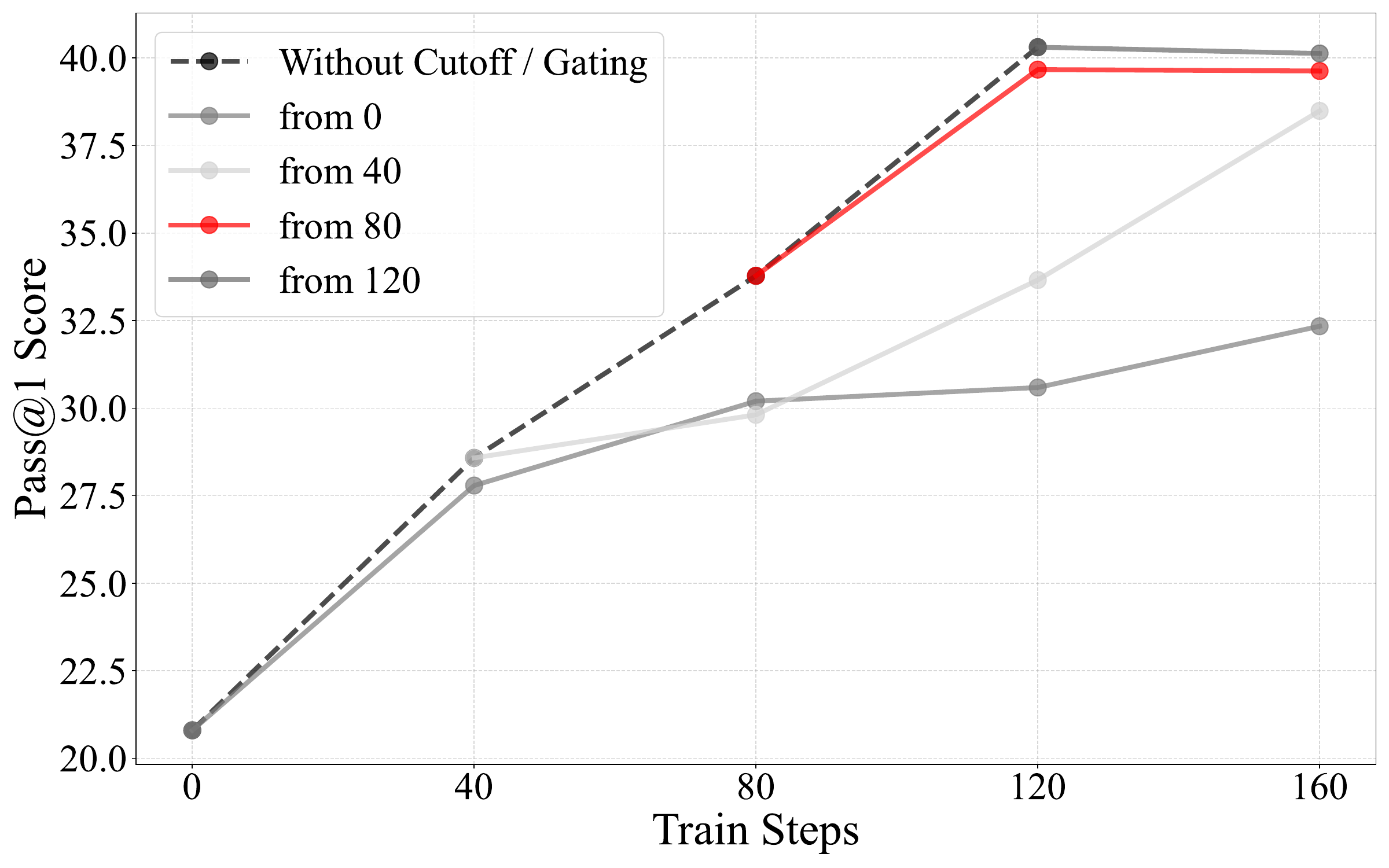}
        \caption{Start Step Ablation.}
        \label{fig:start_step}
    \end{subfigure}
    \caption{\textbf{Ablation on Hyper-parameters for \effname{}.} Pass@1 scores over choices of (a) predictive gating (PG), (b) length cutoff (LC) and (c) start step (k).}
\end{figure*}

\subsection{Ablation Studies}\label{subsec:ablation}

\paragraph{Meta-prediction Components}
To attribute performance improvements to individual factors, we employ a Shapley-$R^2$ decomposition based on linear regression and plot the value in \Cref{fig:Shapley}. We let design matrix as the feature matrix composed of paired differences $(\Delta r_{\text{difficulty}}, \Delta r_{\text{length}}, \Delta r_{\text{notion}}, \Delta \text{step})$, and let $\Delta \text{Acc}$ be the target variable to compute the Shapley-$R^2$ values. The details are deferred due to a spatial constraint.

Moreover, we conduct ablation on training our method \metaname{} with each of the three components and using all three. The results in \cref{fig:radar} show that using all three components of meta-prediction shows overall superior performance over all benchmarks.

\paragraph{Number of Meta Rollouts}

In \Cref{fig:n_rollout}, we analyze the effect of reducing the number of meta-prediction path from 16, which is the default rollout number for the primary solution path. Evaluation shows that using 8 rollouts for meta-prediction, in combination with 16 rollouts for a solution path, shows that optimal result in terms of both train compute and performance.

Introducing the meta-prediction path requires the policy model to generate additional meta-predictions on the solution length, pass-rate, and high-level concepts. However, we show that the average token length and number of rollouts additionally required for meta-predictions only amount to 15.5\% of total rollout compute as shown in \cref{tab:additional_compute}.

\begin{table}[h]
    \centering
    \caption{Comparison on token length between original rollouts and meta prediction rollouts. The increased token is only 15.5\% of entire tokens.}
    \resizebox{0.8\linewidth}{!}{
        \begin{tabular}{l|ccc}
        \hline
        \rowhighlight
        & Avg. Tokens & Rollout No. & Proportion \\
        \hline
        Solution  & 6251 & 16 & 84.5\% \\
        Meta-Pred & 2293 & 8  & 15.5\% \\
        \bottomrule
        \end{tabular}
    }
    \label{tab:additional_compute}
\end{table}

\paragraph{Hyper-parameters for \effname{} ($\sigma_\mathrm{PG}$, $l_\mathrm{LC}$)}
Following an initial $k$-step training phase for length and difficulty meta-prediction, \effname{} applies predictive gating ($\sigma_\mathrm{PG}$) and length cutoff ($l_\mathrm{LC}$).

In \cref{fig:pg}, we evaluate the impact of the predictive gating parameter, $\sigma_\mathrm{PG}$, which determines the threshold for skipping prompts based on the standard deviation of predicted difficulty. Lower values of $\sigma_\mathrm{PG}$ ensure that gating only occurs when the meta-predictions have low variance regarding task difficulty. Similarly, \cref{fig:lc} illustrates the effect of the length cutoff margin, $l_\mathrm{LC}$, a multiplier that scales the threshold for early trace termination. While both PG and LC incur a marginal performance trade-off relative to \metaname{}, these costs are effectively offset by substantial gains in training efficiency and reduced wall-clock time.

\paragraph{Optimizing the Transition Step $k$} The timing of these efficiency mechanisms is critical, as the \effname{} requires an initial calibration period. \cref{fig:start_step} compares the performance of four start-step variants ($k \in \{0, 40, 80, 120\}$) across 160 training steps. We observe that while premature activation ($k < 80$) slightly degrades final accuracy, initiating predictive gating and length cutoff at step 80 achieves performance parity with later starts (e.g., $k=120$) while providing earlier computational savings. We set $k=80$ for \effname{}, as it represents the optimal balance between meta-prediction calibration and resource efficiency. We show that this result is consistent across model sizes in \cref{tab:qwen14b_start_step}. All configurations converge to consistent final performance even when starting from different start steps for 14B model.

\begin{table}[b]
\centering
\caption{Transitioning step ablation on Qwen3-14B Base Model. }
\label{tab:qwen14b_start_step}

\resizebox{0.8\linewidth}{!}{
\begin{tabular}{c|cccc}
\hline
\rowhighlight
Start Step & AIME24 & AIME25 & AMC23 & Avg \\
\hline
0   & 35.83 & 26.04 & 75.78 & 45.88 \\
40  & 33.12 & 25.83 & 75.31 & 44.75 \\
80  & 34.27 & 26.98 & 75.94 & 45.73 \\
120 & 30.52 & 27.40 & 77.34 & 45.09 \\
\hline
\end{tabular}
}

\end{table}

\paragraph{Base Number}
The choice of base number for difficulty reward $r_\text{difficulty}$ is set as 0.01 to  halve the reward per unit difference between predicted and true difficulty. We test the robustness of our method \metaname{} on difference base numbers 0.05, 0.01, and 0.02 on 4B and 8B scale models. As shown in \cref{tab:base_number_ablation}, the results with different base numbers show consistent scores across model sizes and base numbers, except for extreme value of 0.005 on the 4B model, which degrades performance.

\begin{table}[t]
\small
\setlength{\extrarowheight}{3pt}
\centering
\caption{Performance comparison across different base numbers for 4B and 8B models.}
\label{tab:base_number_ablation}

\begin{subtable}[b]{\linewidth}
\centering
\caption{\textbf{Qwen3-4B Base Model}}
\resizebox{\linewidth}{!}{
\begin{tabular}{c|ccccccc}
\hline
\rowhighlight
Base Num & \textbf{AIME'24} & \textbf{AIME'25} & \textbf{AMC'23} & \textbf{MATH500} & \textbf{Minerva} & \textbf{Olympiad} & \textbf{Avg} \\
\midrule
0.005 & 16.98 & 14.58 & 61.95 & 79.62 & \textbf{43.55} & 45.21 & 43.65 \\
0.01  & 26.15 & 21.56 & 70.16 & 84.52 & 41.12 & \textbf{53.38} & 49.48 \\
0.02  & \textbf{26.77} & \textbf{23.33} & \textbf{70.47} & \textbf{84.84} & 43.11 & 53.24 & \textbf{50.29} \\
\bottomrule
\end{tabular}
}
\end{subtable}

\vspace{0.6em}

\begin{subtable}[b]{\linewidth}
\centering
\caption{\textbf{Qwen3-8B Base Model}}
\resizebox{\linewidth}{!}{
\begin{tabular}{c|ccccccc}
\hline
\rowhighlight
Base Num & \textbf{AIME'24} & \textbf{AIME'25} & \textbf{AMC'23} & \textbf{MATH500} & \textbf{Minerva} & \textbf{Olympiad} & \textbf{Avg} \\
\midrule
0.005 & \textbf{34.38} & 25.73 & 78.05 & \textbf{88.34} & \textbf{48.35} & \textbf{57.61} & 55.41 \\
0.01  & 34.17 & \textbf{28.44} & 79.53 & 88.05 & 47.21 & 56.86 & \textbf{55.71} \\
0.02  & 33.54 & 24.17 & \textbf{79.92} & {87.74} & 46.19 & 56.51 & 54.68 \\
\bottomrule
\end{tabular}
}
\end{subtable}
\end{table}

\begin{table}[t]
    \small
    \setlength{\extrarowheight}{3pt}
    \setlength{\tabcolsep}{12pt}
    \centering
    \caption{\textbf{Performance of \metaname{} on Qwen3-8B across six mathematical benchmarks.}  
    All metrics are \textbf{Pass@1}. NF denotes Notion-FeedIn.}
    \label{tab:notion-feedin}
    \resizebox{\linewidth}{!}{
    \begin{tabular}{l|c|c|c|c}
    \rowhighlight
    \hline
        & \textbf{GRPO} 
        & \textbf{GRPO + NF} 
        & \textbf{\metaname{}} 
        & \textbf{\metaname{} + NF} \\
    \hline
    AIME'24  & 28.54\scriptsize{$\pm$4.12} & \textbf{33.96}\scriptsize{$\pm$5.88} & 34.17\scriptsize{$\pm$5.54} & \textbf{35.10\scriptsize{$\pm$4.96}} \\
    AIME'25  & 22.19\scriptsize{$\pm$3.63} & \textbf{23.85}\scriptsize{$\pm$3.64} & \textbf{28.44\scriptsize{$\pm$5.41}} & 25.94\scriptsize{$\pm$4.34} \\
    AMC'23   & 73.67\scriptsize{$\pm$5.60} & \textbf{77.97}\scriptsize{$\pm$5.08} & \textbf{79.53\scriptsize{$\pm$4.26}} & 78.91\scriptsize{$\pm$5.01} \\
    MATH500  & 85.75\scriptsize{$\pm$0.66} & \textbf{86.52}\scriptsize{$\pm$1.14} & 88.05\scriptsize{$\pm$0.82} & \textbf{88.51\scriptsize{$\pm$0.79}} \\
    Minerva  & 43.21\scriptsize{$\pm$2.12} & \textbf{45.44}\scriptsize{$\pm$1.54} & 47.21\scriptsize{$\pm$1.74} & \textbf{48.38\scriptsize{$\pm$1.33}} \\ 
    Olympiad & 54.03\scriptsize{$\pm$1.22} & \textbf{56.63}\scriptsize{$\pm$1.10} & 56.86\scriptsize{$\pm$0.85} & \textbf{57.06\scriptsize{$\pm$0.97}} \\
    \bottomrule
    \end{tabular}
    }
\end{table}

\begin{table}[t]
\small
\setlength{\extrarowheight}{3pt}
\centering
\caption{Performance comparison of \metaname{} with DAPO, trained with Qwen3-8B base model.}
\label{tab:with_dapo}
\resizebox{\linewidth}{!}{
\begin{tabular}{lcc|cc}
\hline
\rowhighlight
& \multicolumn{2}{c}{\textbf{DAPO}} & \multicolumn{2}{c}{\textbf{DAPO + \metaname{}}} \\
\hline
\textbf{Benchmark} & Pass@1 & Pass@8 & Pass@1 & Pass@8 \\
\hline
AIME'24   & 29.48\scriptsize{$\pm$4.04} & 52.54\scriptsize{$\pm$3.99} & \textbf{36.56\scriptsize{$\pm$5.97}} {\color{cadmiumgreen}{(+ 24.02\%)}} & \textbf{66.28\scriptsize{$\pm$3.67}} {\color{cadmiumgreen}{(+ 26.15\%)}}\\
AIME'25   & 23.75\scriptsize{$\pm$3.83} & 37.00\scriptsize{$\pm$2.96} & \textbf{25.94\scriptsize{$\pm$4.39}} {\color{cadmiumgreen}{(+ 9.22\%)}} & \textbf{42.04\scriptsize{$\pm$2.83}} {\color{cadmiumgreen}{(+ 13.62\%)}} \\
AMC'23    & 78.12\scriptsize{$\pm$5.06} & 94.86\scriptsize{$\pm$1.98} & \textbf{78.52\scriptsize{$\pm$4.56}} {\color{cadmiumgreen}{(+ 0.51\%)}}  & \textbf{95.14\scriptsize{$\pm$1.97}} {\color{cadmiumgreen}{(+ 0.29\%)}} \\
MATH500   & 87.44\scriptsize{$\pm$0.74} & 94.40\scriptsize{$\pm$0.38} & \textbf{88.96\scriptsize{$\pm$0.85}} {\color{cadmiumgreen}{(+ 1.74\%)}} & \textbf{95.10\scriptsize{$\pm$0.39}} {\color{cadmiumgreen}{(+ 0.74\%)}} \\
Minerva   & 45.22\scriptsize{$\pm$1.80} & 65.43\scriptsize{$\pm$1.04} & \textbf{47.97\scriptsize{$\pm$2.10}} {\color{cadmiumgreen}{(+ 6.08\%)}} & \textbf{68.77\scriptsize{$\pm$0.99}} {\color{cadmiumgreen}{(+ 5.10\%)}} \\
Olympiad  & 55.97\scriptsize{$\pm$0.89} & 71.13\scriptsize{$\pm$0.63} & \textbf{57.53\scriptsize{$\pm$1.08}} {\color{cadmiumgreen}{(+ 2.79\%)}} & \textbf{73.61\scriptsize{$\pm$0.73}} {\color{cadmiumgreen}{(+ 3.49\%)}} \\
\hline
\textbf{Average} & 53.33\scriptsize{$\pm2.73$} & 69.23\scriptsize{$\pm1.83$} & \textbf{55.01\scriptsize{$\pm$3.16}} {\color{cadmiumgreen}{(+ 3.15\%)}} & \textbf{73.49\scriptsize{$\pm$1.76}} {\color{cadmiumgreen}{(+ 6.15\%)}} \\
\bottomrule
\end{tabular}
}
\end{table}

\begin{table}[t]
\small
\setlength{\extrarowheight}{3pt}
\centering
\caption{Comparative performance of \metaname{} and GRPO across model variants.}
\label{tab:combined_results}

\begin{subtable}[b]{\linewidth}
    \centering
    \caption{Llama 3.1 8B Instruct (3 Epochs / 174 steps)}
    \resizebox{\linewidth}{!}{
    \begin{tabular}{lcc|cc}
    \hline
    \rowhighlight
    & \multicolumn{2}{c}{\textbf{GRPO}} & \multicolumn{2}{c}{\metaname{}} \\
    \hline
    \textbf{Benchmark} & Pass@1 & Pass@8 & Pass@1 & Pass@8 \\
    \hline
    AMC'23      & 25.23\scriptsize{$\pm$4.16} & 42.92\scriptsize{$\pm$3.23} & \textbf{31.02\scriptsize{$\pm$3.88}} & \textbf{56.52\scriptsize{$\pm$3.42}} \\
    Math500     & 52.80\scriptsize{$\pm$1.47} & 71.27\scriptsize{$\pm$0.91} & \textbf{53.54\scriptsize{$\pm$1.14}} & \textbf{71.68\scriptsize{$\pm$0.79}} \\
    Minerva     & 31.70\scriptsize{$\pm$1.69} & 50.15\scriptsize{$\pm$1.21} & \textbf{31.86\scriptsize{$\pm$1.54}} & \textbf{50.39\scriptsize{$\pm$1.19}} \\
    Olympiad    & 19.39\scriptsize{$\pm$0.87} & 32.36\scriptsize{$\pm$0.66} & \textbf{19.89\scriptsize{$\pm$0.79}} & \textbf{35.61\scriptsize{$\pm$0.70}} \\
    \hline
    \end{tabular}
    }
\end{subtable}
\hfill
\begin{subtable}[b]{\linewidth}
    \centering
    \caption{Gemma 2 9B IT (3 Epochs / 174 steps)}
    \resizebox{\linewidth}{!}{
    \begin{tabular}{lcc|cc}
    \hline
    \rowhighlight
    & \multicolumn{2}{c}{\textbf{GRPO}} & \multicolumn{2}{c}{\metaname{}} \\
    \hline
    \textbf{Benchmark} & Pass@1 & Pass@8 & Pass@1 & Pass@8 \\
    \hline
    AMC'23      & 26.88\scriptsize{$\pm$4.80} & 46.48\scriptsize{$\pm$3.84} & \textbf{29.22\scriptsize{$\pm$4.52}} & \textbf{58.88\scriptsize{$\pm$3.54}} \\
    Math500     & 54.07\scriptsize{$\pm$0.90} & 71.70\scriptsize{$\pm$0.79} & \textbf{57.24\scriptsize{$\pm$1.16}} & \textbf{76.57\scriptsize{$\pm$0.75}} \\
    Minerva     & \textbf{34.09\scriptsize{$\pm$1.54}} & 47.90\scriptsize{$\pm$0.98} & 33.58\scriptsize{$\pm$1.85} & \textbf{49.57\scriptsize{$\pm$1.11}} \\
    Olympiad    & 20.29\scriptsize{$\pm$0.95} & 36.56\scriptsize{$\pm$0.75} & \textbf{21.66\scriptsize{$\pm$0.80}} & \textbf{38.59\scriptsize{$\pm$0.76}} \\
    \hline
    \end{tabular}
    }
\end{subtable}
\end{table}
\paragraph{Meta-Predicted Notion Feedin}
We test whether the notions generated from meta-predictions serve as a auxiliary hint for the original solution rollouts by incorporating it into the question using prompt format: Question + `The problem could be solved using following math notions'.
To examine the impact of such notion feed-in on performance, we conduct the following experiment which uses the predicted notions as hints to the question solving phase.

As shown in \Cref{tab:notion-feedin}, incorporating notion feed-in (\metaname{} + NF) yields a small amount of performance gain compared to the variant without notion feed-in (\metaname{}), but the gain is limited. This suggests a high degree of information overlap, implying that the model likely implicitly possesses these concepts through \metaname{}, which enhances meta-awareness, making explicit hinting redundant. Therefore, we test whether the extracted notions boost the performance of a baseline GRPO model. The notions are extracted from the MAPR model and fed into a separately trained GRPO model using keywords with high notion reward scores. Although this setting is far from practical deployment, as it requires cross-model notion extraction and transfer, the substantial improvement achieved by GRPO + NF demonstrates that the extracted notions are highly effective in enhancing reasoning performance.

\paragraph{Ablation on RL Algorithm}
\metaname{} is flexibly applicable to GRPO variants. We show the superiority of \metaname{} combined with DAPO algorithm in \Cref{tab:with_dapo}.
Unlike DAPO, which requires a redundant sampling phase to filter out tasks with zero-variance, our method is able to bypass the sampling for solution rollouts and preemptively gate such tasks. Even with greater efficiency, \metaname{} outperforms all six mathematical benchmarks by a large margin. 

We train Qwen3-8B-Base with DAPO for three epochs (315 steps), which is equivalent to one epoch of GRPO (314 steps) in terms of the total number of gradient updates. Moreover, we disable the overlong reward shaping term in DAPO. In our setting, this term imposes an overly strong length constraint, which prevents the model from sufficiently increasing its reasoning depth. Empirically, we observe that keeping this term results in lower final performance. We therefore remove it to avoid unnecessarily restricting the model’s reasoning capacity under our training configuration.

\paragraph{Ablation on Different Model}
Our method also demonstrates consistent improvements when applied to different model families, Llama 3.1 8B Instruct~\citep{grattafiori2024llama} and Gemma 2 9B IT~\citep{team2024gemma}. Unlike Qwen3 family that are explicitly trained on long CoT reasoning datset, these two model families are relatively under-trained on mathematical reasoning dataset. Therefore, following the convention of existing works~\citep{zhu2025the,liu2025understanding}, we train both models with easier dataset, train split of MATH dataset, with extended training epochs of 3. All the other configurations are kept the same. In \Cref{tab:combined_results}, we report the evaluation result on AMC’23, MATH500, Minerva, and OlympiadBench, excluding AIME’24 and AIME’25 for extremely low accuracy nearing 0 for both methods even after training. Overall, our method achieves consistent gains not only for Qwen3 but also for Llama 3.1 and Gemma 2 models.

\section*{Conclusion}
We present \metaname{}, a meta-aware reinforcement learning framework that fosters meta-cognitive ability by self-alignment. By incorporating information achieved by meta-thinking trajectories into training, our method enables stable and efficient optimization by integrating predictive gating and early cutoff. Empirically, \metaname{} accelerates RL-based post-training while improving both in-domain and out-of-domain performance, demonstrating notable gains in accuracy and generalization. These results highlight the promise of meta prediction as a principled avenue for enhancing reasoning models.

\section*{Impact Statement}
This paper presents \metaname{}, a framework designed to verify and utilize meta-awareness in reasoning models. The primary broader impact of our work lies in the improvement of computational efficiency for Large Language Models. By enabling models to self-regulate, such as determining optimal thinking duration and filtering out unsolvable prompts, our approach significantly reduces the computational resources required for both training and inference. This contributes to reducing the environmental footprint associated with developing and deploying large-scale reasoning systems. While advancing reasoning capabilities generally implies the need for careful consideration of dual-use risks, our work specifically focuses on internal verification and efficiency, and we do not foresee specific negative societal consequences unique to this method.

\clearpage
\section*{Acknowledgement}
This work was supported by Institute for Information \& communications Technology Planning \& Evaluation(IITP) grant funded by the Korea government(MSIT) (RS-2019-II190075, Artificial Intelligence Graduate School Program(KAIST)) and National Research Foundation of Korea (NRF) grant (No.RS-2023-00209060, A Study on Optimization and Network Interpretation Method for Large-Scale Machine Learning) funded by the Korea government (MSIT).

\bibliography{bib/custom}

@article{deepseekmath_grpo,
  title        = {DeepSeekMath: Pushing the Limits of Mathematical Reasoning in Open Language Models},
  author       = {Zhihong Shao and Peiyi Wang and Qihao Zhu and Runxin Xu and Junxiao Song and Xiao Bi and Haowei Zhang and Mingchuan Zhang and Y. K. Li and Y. Wu and Daya Guo},
  year         = {2024},
  archivePrefix= {arXiv},
  eprint       = {2402.03300},
  primaryClass = {cs.CL},
  doi          = {10.48550/arXiv.2402.03300},
  url          = {https://arxiv.org/abs/2402.03300}
}

@article{lcpo_l1,
  title        = {L1: Controlling How Long A Reasoning Model Thinks With Reinforcement Learning},
  author       = {Pranjal Aggarwal and Sean Welleck},
  year         = {2025},
  archivePrefix= {arXiv},
  eprint       = {2503.04697},
  primaryClass = {cs.CL},
  doi          = {10.48550/arXiv.2503.04697},
  url          = {https://arxiv.org/abs/2503.04697}
}

@article{lmpo,
  title        = {Length-Controlled Margin-Based Preference Optimization without Reference Model},
  author       = {Gengxu Li and Tingyu Xia and Yi Chang and Yuan Wu},
  year         = {2025},
  archivePrefix= {arXiv},
  eprint       = {2502.14643},
  primaryClass = {cs.CL},
  doi          = {10.48550/arXiv.2502.14643},
  url          = {https://arxiv.org/abs/2502.14643}
}

@article{alp,
  title        = {Just Enough Thinking: Efficient Reasoning with Adaptive Length Penalties Reinforcement Learning},
  author       = {Violet Xiang and Chase Blagden and Rafael Rafailov and Nathan Lile and Sang Truong and Chelsea Finn and Nick Haber},
  year         = {2025},
  archivePrefix= {arXiv},
  eprint       = {2506.05256},
  primaryClass = {cs.AI},
  doi          = {10.48550/arXiv.2506.05256},
  url          = {https://arxiv.org/abs/2506.05256}
}

@article{adactrl,
  title        = {AdaCtrl: Towards Adaptive and Controllable Reasoning via Difficulty-Aware Budgeting},
  author       = {Shijue Huang and Hongru Wang and Wanjun Zhong and Zhaochen Su and Jiazhan Feng and Bowen Cao and Yi R. Fung},
  year         = {2025},
  archivePrefix= {arXiv},
  eprint       = {2505.18822},
  primaryClass = {cs.AI},
  doi          = {10.48550/arXiv.2505.18822},
  url          = {https://arxiv.org/abs/2505.18822}
}

@article{grpo_lead,
  title        = {GRPO-LEAD: A Difficulty-Aware Reinforcement Learning Approach for Concise Mathematical Reasoning in Language Models},
  author       = {Jixiao Zhang and Chunsheng Zuo},
  year         = {2025},
  archivePrefix= {arXiv},
  eprint       = {2504.09696},
  primaryClass = {cs.CL},
  doi          = {10.48550/arXiv.2504.09696},
  url          = {https://arxiv.org/abs/2504.09696}
}

@article{mera,
  title        = {From "Aha Moments" to Controllable Thinking: Toward Meta-Cognitive Reasoning in Large Reasoning Models via Decoupled Reasoning and Control},
  author       = {Rui Ha and Chaozhuo Li and Rui Pu and Sen Su},
  year         = {2025},
  archivePrefix= {arXiv},
  eprint       = {2508.04460},
  primaryClass = {cs.AI},
  doi          = {10.48550/arXiv.2508.04460},
  url          = {https://arxiv.org/abs/2508.04460}
}

@article{rlvmr,
  title        = {RLVMR: Reinforcement Learning with Verifiable Meta-Reasoning Rewards for Robust Long-Horizon Agents},
  author       = {Zijing Zhang and Ziyang Chen and Mingxiao Li and Zhaopeng Tu and Xiaolong Li},
  year         = {2025},
  archivePrefix= {arXiv},
  eprint       = {2507.22844},
  primaryClass = {cs.LG},
  doi          = {10.48550/arXiv.2507.22844},
  url          = {https://arxiv.org/abs/2507.22844}
}

@article{difficultystage,
  title        = {How Difficulty-Aware Staged Reinforcement Learning Enhances LLMs' Reasoning Capabilities: A Preliminary Experimental Study},
  author       = {Yunjie Ji and Sitong Zhao and Xiaoyu Tian and Haotian Wang and Shuaiting Chen and Yiping Peng and Han Zhao and Xiangang Li},
  year         = {2025},
  archivePrefix= {arXiv},
  eprint       = {2504.00829},
  primaryClass = {cs.AI},
  doi          = {10.48550/arXiv.2504.00829},
  url          = {https://arxiv.org/abs/2504.00829}
}

@article{preview_difficult_intervention,
  title        = {Enhancing Math Reasoning in Small-sized LLMs via Preview Difficulty-Aware Intervention},
  author       = {Xinhan Di and JoyJiaoW},
  year         = {2025},
  archivePrefix= {arXiv},
  eprint       = {2508.01604},
  primaryClass = {cs.LG},
  doi          = {10.48550/arXiv.2508.01604},
  url          = {https://arxiv.org/abs/2508.01604}
}

@article{drgrpo,
  title        = {Understanding R1-Zero-Like Training: A Critical Perspective},
  author       = {Zichen Liu and Changyu Chen and Wenjun Li and Penghui Qi and Tianyu Pang and Chao Du and Wee Sun Lee and Min Lin},
  year         = {2025},
  archivePrefix= {arXiv},
  eprint       = {2503.20783},
  primaryClass = {cs.LG},
  doi          = {10.48550/arXiv.2503.20783},
  url          = {https://arxiv.org/abs/2503.20783}
}

@article{acereason_nemotron,
  title        = {AceReason-Nemotron: Advancing Math and Code Reasoning through Reinforcement Learning},
  author       = {Yang Chen and Zhuolin Yang and Zihan Liu and Chankyu Lee and Peng Xu and Mohammad Shoeybi and Bryan Catanzaro and Wei Ping},
  year         = {2025},
  archivePrefix= {arXiv},
  eprint       = {2505.16400},
  primaryClass = {cs.CL},
  doi          = {10.48550/arXiv.2505.16400},
  url          = {https://arxiv.org/abs/2505.16400}
}

@article{zheng2025group,
  title={Group sequence policy optimization},
  author={Zheng, Chujie and Liu, Shixuan and Li, Mingze and Chen, Xiong-Hui and Yu, Bowen and Gao, Chang and Dang, Kai and Liu, Yuqiong and Men, Rui and Yang, An and others},
  journal={arXiv preprint arXiv:2507.18071},
  year={2025}
}

@article{yu2025dapo,
  title={Dapo: An open-source llm reinforcement learning system at scale},
  author={Yu, Qiying and Zhang, Zheng and Zhu, Ruofei and Yuan, Yufeng and Zuo, Xiaochen and Yue, Yu and Dai, Weinan and Fan, Tiantian and Liu, Gaohong and Liu, Lingjun and others},
  journal={arXiv preprint arXiv:2503.14476},
  year={2025}
}

@article{brown2020language,
  title={Language models are few-shot learners},
  author={Brown, Tom and Mann, Benjamin and Ryder, Nick and Subbiah, Melanie and Kaplan, Jared D and Dhariwal, Prafulla and Neelakantan, Arvind and Shyam, Pranav and Sastry, Girish and Askell, Amanda and others},
  journal={Advances in neural information processing systems},
  volume={33},
  pages={1877--1901},
  year={2020}
}

@article{touvron2023llama,
  title={Llama: Open and efficient foundation language models},
  author={Touvron, Hugo and Lavril, Thibaut and Izacard, Gautier and Martinet, Xavier and Lachaux, Marie-Anne and Lacroix, Timoth{\'e}e and Rozi{\`e}re, Baptiste and Goyal, Naman and Hambro, Eric and Azhar, Faisal and others},
  journal={arXiv preprint arXiv:2302.13971},
  year={2023}
}

@article{yang2025qwen3,
  title={Qwen3 technical report},
  author={Yang, An and Li, Anfeng and Yang, Baosong and Zhang, Beichen and Hui, Binyuan and Zheng, Bo and Yu, Bowen and Gao, Chang and Huang, Chengen and Lv, Chenxu and others},
  journal={arXiv preprint arXiv:2505.09388},
  year={2025}
}

@article{guo2025deepseek,
  title={Deepseek-r1: Incentivizing reasoning capability in llms via reinforcement learning},
  author={Guo, Daya and Yang, Dejian and Zhang, Haowei and Song, Junxiao and Zhang, Ruoyu and Xu, Runxin and Zhu, Qihao and Ma, Shirong and Wang, Peiyi and Bi, Xiao and others},
  journal={arXiv preprint arXiv:2501.12948},
  year={2025}
}

@article{dai2025s,
  title={S-GRPO: Early Exit via Reinforcement Learning in Reasoning Models},
  author={Dai, Muzhi and Yang, Chenxu and Si, Qingyi},
  journal={arXiv preprint arXiv:2505.07686},
  year={2025}
}

@article{han2024token,
  title={Token-budget-aware llm reasoning},
  author={Han, Tingxu and Wang, Zhenting and Fang, Chunrong and Zhao, Shiyu and Ma, Shiqing and Chen, Zhenyu},
  journal={arXiv preprint arXiv:2412.18547},
  year={2024}
}

@article{yang2025think,
  title={Think when you need: Self-adaptive chain-of-thought learning},
  author={Yang, Junjie and Lin, Ke and Yu, Xing},
  journal={arXiv preprint arXiv:2504.03234},
  year={2025}
}

@article{fang2025thinkless,
  title={Thinkless: Llm learns when to think},
  author={Fang, Gongfan and Ma, Xinyin and Wang, Xinchao},
  journal={arXiv preprint arXiv:2505.13379},
  year={2025}
}

@article{zhang2025edge,
  title={EDGE-GRPO: Entropy-Driven GRPO with Guided Error Correction for Advantage Diversity},
  author={Zhang, Xingjian and Wen, Siwei and Wu, Wenjun and Huang, Lei},
  journal={arXiv preprint arXiv:2507.21848},
  year={2025}
}

@article{he2025good,
  title={Good Learners Think Their Thinking: Generative PRM Makes Large Reasoning Model More Efficient Math Learner},
  author={He, Tao and Mu, Rongchuan and Liao, Lizi and Cao, Yixin and Liu, Ming and Qin, Bing},
  journal={arXiv preprint arXiv:2507.23317},
  year={2025}
}

@article{sui2025meta,
  title={Meta-reasoner: Dynamic guidance for optimized inference-time reasoning in large language models},
  author={Sui, Yuan and He, Yufei and Cao, Tri and Han, Simeng and Chen, Yulin and Hooi, Bryan},
  journal={arXiv preprint arXiv:2502.19918},
  year={2025}
}

@article{wang2025adaptive,
  title={Adaptive Deep Reasoning: Triggering Deep Thinking When Needed},
  author={Wang, Yunhao and Zhang, Yuhao and Yu, Tinghao and Xu, Can and Zhang, Feng and Lian, Fengzong},
  journal={arXiv preprint arXiv:2505.20101},
  year={2025}
}

@article{chen2025aware,
  title={Aware First, Think Less: Dynamic Boundary Self-Awareness Drives Extreme Reasoning Efficiency in Large Language Models},
  author={Chen, Qiguang and Peng, Dengyun and Liu, Jinhao and Su, HuiKang and Guan, Jiannan and Qin, Libo and Che, Wanxiang},
  journal={arXiv preprint arXiv:2508.11582},
  year={2025}
}

@article{yang2025dynamic,
  title={Dynamic Early Exit in Reasoning Models},
  author={Yang, Chenxu and Si, Qingyi and Duan, Yongjie and Zhu, Zheliang and Zhu, Chenyu and Li, Qiaowei and Lin, Zheng and Cao, Li and Wang, Weiping},
  journal={arXiv preprint arXiv:2504.15895},
  year={2025}
}

@article{qiao2025concise,
  title={ConCISE: Confidence-guided Compression in Step-by-step Efficient Reasoning},
  author={Qiao, Ziqing and Deng, Yongheng and Zeng, Jiali and Wang, Dong and Wei, Lai and Meng, Fandong and Zhou, Jie and Ren, Ju and Zhang, Yaoxue},
  journal={arXiv preprint arXiv:2505.04881},
  year={2025}
}

@article{lu2025prolonged,
  title={Prolonged reasoning is not all you need: Certainty-based adaptive routing for efficient llm/mllm reasoning},
  author={Lu, Jinghui and Yu, Haiyang and Xu, Siliang and Ran, Shiwei and Tang, Guozhi and Wang, Siqi and Shan, Bin and Fu, Teng and Feng, Hao and Tang, Jingqun and others},
  journal={arXiv preprint arXiv:2505.15154},
  year={2025}
}

@article{zhang2025adaptthink,
  title={Adaptthink: Reasoning models can learn when to think},
  author={Zhang, Jiajie and Lin, Nianyi and Hou, Lei and Feng, Ling and Li, Juanzi},
  journal={arXiv preprint arXiv:2505.13417},
  year={2025}
}

@article{shen2025dast,
  title={Dast: Difficulty-adaptive slow-thinking for large reasoning models},
  author={Shen, Yi and Zhang, Jian and Huang, Jieyun and Shi, Shuming and Zhang, Wenjing and Yan, Jiangze and Wang, Ning and Wang, Kai and Liu, Zhaoxiang and Lian, Shiguo},
  journal={arXiv preprint arXiv:2503.04472},
  year={2025}
}

@article{qu2025optimizing,
  title={Optimizing test-time compute via meta reinforcement fine-tuning},
  author={Qu, Yuxiao and Yang, Matthew YR and Setlur, Amrith and Tunstall, Lewis and Beeching, Edward Emanuel and Salakhutdinov, Ruslan and Kumar, Aviral},
  journal={arXiv preprint arXiv:2503.07572},
  year={2025}
}

@article{tu2025learning,
  title={Learning When to Think: Shaping Adaptive Reasoning in R1-Style Models via Multi-Stage RL},
  author={Tu, Songjun and Lin, Jiahao and Zhang, Qichao and Tian, Xiangyu and Li, Linjing and Lan, Xiangyuan and Zhao, Dongbin},
  journal={arXiv preprint arXiv:2505.10832},
  year={2025}
}

@article{shi2025efficient,
  title={Efficient reinforcement finetuning via adaptive curriculum learning},
  author={Shi, Taiwei and Wu, Yiyang and Song, Linxin and Zhou, Tianyi and Zhao, Jieyu},
  journal={arXiv preprint arXiv:2504.05520},
  year={2025}
}

@article{liu2025ghpo,
  title={GHPO: Adaptive Guidance for Stable and Efficient LLM Reinforcement Learning},
  author={Liu, Ziru and Gong, Cheng and Fu, Xinyu and Liu, Yaofang and Chen, Ran and Hu, Shoubo and Zhang, Suiyun and Liu, Rui and Zhang, Qingfu and Tu, Dandan},
  journal={arXiv preprint arXiv:2507.10628},
  year={2025}
}

@article{de2024rational,
  title={Rational metareasoning for large language models},
  author={De Sabbata, C Nicol{\`o} and Sumers, Theodore R and AlKhamissi, Badr and Bosselut, Antoine and Griffiths, Thomas L},
  journal={arXiv preprint arXiv:2410.05563},
  year={2024}
}

@inproceedings{adamw,
  title={Decoupled Weight Decay Regularization},
  author={Loshchilov, Ilya and Hutter, Frank},
  booktitle={International Conference on Learning Representations}
}

@inproceedings{
liu2025position,
title={Position: Truly Self-Improving Agents Require Intrinsic Metacognitive Learning},
author={Tennison Liu and Mihaela van der Schaar},
booktitle={Forty-second International Conference on Machine Learning Position Paper Track},
year={2025},
url={https://openreview.net/forum?id=4KhDd0Ozqe}
}

@article{dong2025meta,
  title={Meta-R1: Empowering Large Reasoning Models with Metacognition},
  author={Dong, Haonan and Ye, Haoran and Zhu, Wenhao and Jiang, Kehan and Song, Guojie},
  journal={arXiv preprint arXiv:2508.17291},
  year={2025}
}

@article{didolkar_metacognitive,
title={Metacognitive Reuse: Turning Recurring LLM
Reasoning Into Concise Behaviors},
author={Didolkar, Aniket and Balla, Nicolas and Arora, Sanjeev and Goyal, Anirudh},
journal={arXiv preprint arXiv:2509.13237},
  year={2025}
}

@article{wan2025rema,
  title={Rema: Learning to meta-think for llms with multi-agent reinforcement learning},
  author={Wan, Ziyu and Li, Yunxiang and Wen, Xiaoyu and Song, Yan and Wang, Hanjing and Yang, Linyi and Schmidt, Mark and Wang, Jun and Zhang, Weinan and Hu, Shuyue and others},
  journal={arXiv preprint arXiv:2503.09501},
  year={2025}
}

@article{yang2025learning,
  title={Learning to Deliberate: Meta-policy Collaboration for Agentic LLMs with Multi-agent Reinforcement Learning},
  author={Yang, Wei and Thomason, Jesse},
  journal={arXiv preprint arXiv:2509.03817},
  year={2025}
}

@article{bilal2025meta,
  title={Meta-thinking in llms via multi-agent reinforcement learning: A survey},
  author={Bilal, Ahsan and Mohsin, Muhammad Ahmed and Umer, Muhammad and Bangash, Muhammad Awais Khan and Jamshed, Muhammad Ali},
  journal={arXiv preprint arXiv:2504.14520},
  year={2025}
}

@article{khandelwal2025language,
  title={Language Models Coupled with Metacognition Can Outperform Reasoning Models},
  author={Khandelwal, Vedant and Rossi, Francesca and Murugesan, Keerthiram and Miehling, Erik and Campbell, Murray and Ramamurthy, Karthikeyan Natesan and Horesh, Lior},
  journal={arXiv preprint arXiv:2508.17959},
  year={2025}
}

@inproceedings{math500,
  title={Measuring Mathematical Problem Solving With the MATH Dataset},
  author={Hendrycks, Dan and Burns, Collin and Kadavath, Saurav and Arora, Akul and Basart, Steven and Tang, Eric and Song, Dawn and Steinhardt, Jacob},
  booktitle={Thirty-fifth Conference on Neural Information Processing Systems Datasets and Benchmarks Track (Round 2)}
}

@inproceedings{olympiadbench,
  title={OlympiadBench: A Challenging Benchmark for Promoting AGI with Olympiad-Level Bilingual Multimodal Scientific Problems},
  author={He, Chaoqun and Luo, Renjie and Bai, Yuzhuo and Hu, Shengding and Thai, Zhen and Shen, Junhao and Hu, Jinyi and Han, Xu and Huang, Yujie and Zhang, Yuxiang and others},
  booktitle={Proceedings of the 62nd Annual Meeting of the Association for Computational Linguistics (Volume 1: Long Papers)},
  pages={3828--3850},
  year={2024}
}

@article{qiu2025mela,
  title={Mela: A metacognitive llm-driven architecture for automatic heuristic design},
  author={Qiu, Zishang and Chen, Xinan and Chen, Long and Bai, Ruibin},
  journal={arXiv preprint arXiv:2507.20541},
  year={2025}
}

@inproceedings{logicllm,
  title={Logic-LM: Empowering Large Language Models with Symbolic Solvers for Faithful Logical Reasoning},
  author={Pan, Liangming and Albalak, Alon and Wang, Xinyi and Wang, William},
  booktitle={Findings of the Association for Computational Linguistics: EMNLP 2023},
  pages={3806--3824},
  year={2023}
}

@inproceedings{prontoqa,
  title={Language Models Are Greedy Reasoners: A Systematic Formal Analysis of Chain-of-Thought},
  author={Saparov, Abulhair and He, He},
  booktitle={The Eleventh International Conference on Learning Representations}
}

@inproceedings{proofwriter,
  title={ProofWriter: Generating Implications, Proofs, and Abductive Statements over Natural Language},
  author={Tafjord, Oyvind and Dalvi, Bhavana and Clark, Peter},
  booktitle={Findings of the Association for Computational Linguistics: ACL-IJCNLP 2021},
  pages={3621--3634},
  year={2021}
}

@inproceedings{folio,
  title={FOLIO: Natural Language Reasoning with First-Order Logic},
  author={Han, Simeng and Schoelkopf, Hailey and Zhao, Yilun and Qi, Zhenting and Riddell, Martin and Zhou, Wenfei and Coady, James and Peng, David and Qiao, Yujie and Benson, Luke and others},
  booktitle={Proceedings of the 2024 Conference on Empirical Methods in Natural Language Processing},
  pages={22017--22031},
  year={2024}
}

@article{ma2025large,
  title={Large Language Models Have Intrinsic Meta-Cognition, but Need a Good Lens},
  author={Ma, Ziyang and Yuan, Qingyue and Wang, Zhenglin and Zhou, Deyu},
  journal={arXiv preprint arXiv:2506.08410},
  year={2025}
}

@article{bigbench,
  title={Beyond the Imitation Game: Quantifying and extrapolating the capabilities of language models},
  author={Srivastava, Aarohi and Rastogi, Abhinav and Rao, Abhishek and Shoeb, Abu Awal Md and Abid, Abubakar and Fisch, Adam and Brown, Adam R and Santoro, Adam and Gupta, Aditya and Garriga-Alonso, Adri{\`a} and others},
  journal={Transactions on Machine Learning Research}
}

@inproceedings{AR-LSAT,
    title = "Analytical Reasoning of Text",
    author = "Zhong, Wanjun  and
      Wang, Siyuan  and
      Tang, Duyu  and
      Xu, Zenan  and
      Guo, Daya  and
      Chen, Yining  and
      Wang, Jiahai  and
      Yin, Jian  and
      Zhou, Ming  and
      Duan, Nan",
    editor = "Carpuat, Marine  and
      de Marneffe, Marie-Catherine  and
      Meza Ruiz, Ivan Vladimir",
    booktitle = "Findings of the Association for Computational Linguistics: NAACL 2022",
    month = jul,
    year = "2022",
    address = "Seattle, United States",
    publisher = "Association for Computational Linguistics",
    url = "https://aclanthology.org/2022.findings-naacl.177/",
    doi = "10.18653/v1/2022.findings-naacl.177",
    pages = "2306--2319"
}

@inproceedings{
rein2024gpqa,
title={{GPQA}: A Graduate-Level Google-Proof Q\&A Benchmark},
author={David Rein and Betty Li Hou and Asa Cooper Stickland and Jackson Petty and Richard Yuanzhe Pang and Julien Dirani and Julian Michael and Samuel R. Bowman},
booktitle={First Conference on Language Modeling},
year={2024},
url={https://openreview.net/forum?id=Ti67584b98}
}

@article{Guo2025RBench,
  title={R-Bench: Graduate-level Multi-disciplinary Benchmarks for LLM \& MLLM Complex Reasoning Evaluation},
  author={Guo, Meng-Hao and Xu, Jiajun and Zhang, Yi and Song, Jiaxi and Peng, Haoyang and Deng, Yi-Xuan and Dong, Xinzhi and Nakayama, Kiyohiro and Geng, Zhengyang and Wang, Chen and Ni, Bolin and Yang, Guo-Wei and Rao, Yongming and Peng, Houwen and Hu, Han and Wetzstein, Gordon and Hu, Shi-min},
  journal={arXiv preprint arXiv:2505.02018},
  year={2025},
  url={https://arxiv.org/abs/2505.02018}
}

@article{Clark2018ARC,
  title={Think you have Solved Question Answering? Try ARC, the AI2 Reasoning Challenge},
  author={Clark, Peter and Cowhey, Isaac and Etzioni, Oren and Khot, Tushar and Sabharwal, Ashish and Schoenick, Carissa and Tafjord, Oyvind},
  journal={arXiv preprint arXiv:1803.05457},
  year={2018},
  url={https://arxiv.org/abs/1803.05457},
  note={Use the ARC-Challenge split for ARC-C results.}
}

@inproceedings{
wang2024scibench,
title={SciBench: Evaluating College-Level Scientific Problem-Solving Abilities of Large Language Models},
author={Xiaoxuan Wang and Ziniu Hu and Pan Lu and Yanqiao Zhu and Jieyu Zhang and Satyen Subramaniam and Arjun R Loomba and Shichang Zhang and Yizhou Sun and Wei Wang},
booktitle={Forty-first International Conference on Machine Learning},
year={2024},
url={https://openreview.net/forum?id=bq1JEgioLr}
}

@article{liu2023your,
  title={Is your code generated by chatgpt really correct? rigorous evaluation of large language models for code generation},
  author={Liu, Jiawei and Xia, Chunqiu Steven and Wang, Yuyao and Zhang, Lingming},
  journal={Advances in Neural Information Processing Systems},
  volume={36},
  pages={21558--21572},
  year={2023}
}

@InProceedings{pmlr-v235-gu24c,
  title = 	 {{CRUXE}val: A Benchmark for Code Reasoning, Understanding and Execution},
  author =       {Gu, Alex and Roziere, Baptiste and Leather, Hugh James and Solar-Lezama, Armando and Synnaeve, Gabriel and Wang, Sida},
  booktitle = 	 {Proceedings of the 41st International Conference on Machine Learning},
  pages = 	 {16568--16621},
  year = 	 {2024},
  editor = 	 {Salakhutdinov, Ruslan and Kolter, Zico and Heller, Katherine and Weller, Adrian and Oliver, Nuria and Scarlett, Jonathan and Berkenkamp, Felix},
  volume = 	 {235},
  series = 	 {Proceedings of Machine Learning Research},
  month = 	 {21--27 Jul},
  publisher =    {PMLR},
  url = 	 {https://proceedings.mlr.press/v235/gu24c.html}
}

@article{austin2021program,
  title={Program synthesis with large language models},
  author={Austin, Jacob and Odena, Augustus and Nye, Maxwell and Bosma, Maarten and Michalewski, Henryk and Dohan, David and Jiang, Ellen and Cai, Carrie and Terry, Michael and Le, Quoc and others},
  journal={arXiv preprint arXiv:2108.07732},
  year={2021}
}

@inproceedings{
jain2025livecodebench,
title={LiveCodeBench: Holistic and Contamination Free Evaluation of Large Language Models for Code},
author={Naman Jain and King Han and Alex Gu and Wen-Ding Li and Fanjia Yan and Tianjun Zhang and Sida Wang and Armando Solar-Lezama and Koushik Sen and Ion Stoica},
booktitle={The Thirteenth International Conference on Learning Representations},
year={2025},
url={https://openreview.net/forum?id=chfJJYC3iL}
}

@misc{deepscaler2025,
  title={DeepScaleR: Surpassing O1-Preview with a 1.5B Model by Scaling RL},
  author={Michael Luo and Sijun Tan and Justin Wong and Xiaoxiang Shi and William Tang and Manan Roongta and Colin Cai and Jeffrey Luo and Tianjun Zhang and Erran Li and Raluca Ada Popa and Ion Stoica},
  year={2025},
  howpublished={\url{https://pretty-radio-b75.notion.site/DeepScaleR-Surpassing-O1-Preview-with-a-1-5B-Model-by-Scaling-RL-19681902c1468005bed8ca303013a4e2}},
  note={Notion Blog}
}

@article{grattafiori2024llama,
  title={The llama 3 herd of models},
  author={Grattafiori, Aaron and Dubey, Abhimanyu and Jauhri, Abhinav and Pandey, Abhinav and Kadian, Abhishek and Al-Dahle, Ahmad and Letman, Aiesha and Mathur, Akhil and Schelten, Alan and Vaughan, Alex and others},
  journal={arXiv preprint arXiv:2407.21783},
  year={2024}
}

@article{team2024gemma,
  title={Gemma 2: Improving open language models at a practical size},
  author={Team, Gemma and Riviere, Morgane and Pathak, Shreya and Sessa, Pier Giuseppe and Hardin, Cassidy and Bhupatiraju, Surya and Hussenot, L{\'e}onard and Mesnard, Thomas and Shahriari, Bobak and Ram{\'e}, Alexandre and others},
  journal={arXiv preprint arXiv:2408.00118},
  year={2024}
}

@inproceedings{liu2025understanding,
  title={Understanding r1-zero-like training: A critical perspective},
  author={Liu, Zichen and Chen, Changyu and Li, Wenjun and Qi, Penghui and Pang, Tianyu and Du, Chao and Lee, Wee Sun and Lin, Min},
  booktitle={Conference on Language Modeling (COLM)},
  year={2025}
}

@inproceedings{
zhu2025the,
title={The Surprising Effectiveness of Negative Reinforcement in {LLM} Reasoning},
author={Xinyu Zhu and Mengzhou Xia and Zhepei Wei and Wei-Lin Chen and Danqi Chen and Yu Meng},
booktitle={The Thirty-ninth Annual Conference on Neural Information Processing Systems},
year={2025},
url={https://openreview.net/forum?id=ftVlLG9cks}
}
\bibliographystyle{icml2026}
\newpage
\appendix
\onecolumn
\clearpage

\section{Default Meta-prediction Prompt for \metaname{}}\label{sec:appendix}

\begin{center}
\begin{myyellowbox}{Prompt}
\textbf{[System]:}

You are a helpful assistant.

\textbf{[User]:} \\
Think step-by-step between $<$meta$>$ and $<$/meta$>$, ensuring comprehensive and detailed reasoning especially for determining the $\mathrm{pass\_rate}$ and $\mathrm{solution\_length}$ values. For each component $(\mathrm{math\_notion}$, $\mathrm{pass\_rate}$, $\mathrm{solution\_length})$, provide a comprehensive illustration or example during your reasoning in the $<$meta$>$ section to clarify how each value is decided. When determining $\mathrm{math\_notion}$, ensure that the notions listed do not directly include the notions already written in the problem statement. After $<$/meta$>$, return a JSON object with three keys:

- $\mathrm{math\_notion}$ (list[str])

- $\mathrm{pass\_rate}$ (integer from 0 to 8)

- $\mathrm{solution\_length}$ (integer from 128 to \{$\mathrm{max\_response\_length}$\})

\vspace{5mm}

Problem: \{problem\}

\end{myyellowbox}
\end{center}

In the meta-prediction prompt, \texttt{math\_notion} is predicted as a \texttt{list[str]}, where each element denotes a mathematical notion required to solve the problem. We avoid predicting a continuous value (or an overly fine-grained scale) for \texttt{pass\_rate}, since it can introduce unnecessary variance and instability in the predicted difficulty. Instead, the prompt restricts \texttt{pass\_rate} to an integer in \{0,\dots,8\}. When computing the reward, we normalize this value by dividing it by 8. Finally, \texttt{solution\_length} is predicted as an integer between 128 and the maximum response length of the corresponding training setup.

\clearpage
\section{Meta-prediction Dynamics During \metaname{} Training}

\paragraph{\metaname{} Performance in Out-of-Domain Benchmarks}

\begin{table*}[h]
\small
\setlength{\extrarowheight}{3pt}
\centering
\caption{\textbf{Performance of GRPO and \metaname{} in Out-of-Domain benchmarks.} Results are reported as pass@1 score.} \label{tab:generalization}
\resizebox{\textwidth}{!}{
\begin{tabular}{l|cc|l|cc|l|cc}
\toprule
\multicolumn{3}{c|}{\textbf{Logical Reasoning}} &
\multicolumn{3}{c|}{\textbf{Scientific Reasoning}} &
\multicolumn{3}{c}{\textbf{Coding}}\\
\cmidrule(lr){1-3}\cmidrule(lr){4-6}\cmidrule(lr){7-9}
\textbf{Benchmark} & \textbf{GRPO} & \textbf{w/ \metaname{}} &
\textbf{Benchmark} & \textbf{GRPO} & \textbf{w/ \metaname{}} &
\textbf{Benchmark} & \textbf{GRPO} & \textbf{w/ \metaname{}} \\
\hline
ProntoQA      & 90.56 & \textbf{93.74} & GPQA Diamond   & 51.72 & \textbf{53.72} & EvalPlus      & 77.32 & \textbf{77.66}\\
ProofWriter   & 72.27 & \textbf{73.23} & R-Bench        & 60.69 & \textbf{61.68} & CRUX-O        & 72.72 & \textbf{73.39}\\
FOLIO         & 69.16 & \textbf{69.24} & ARC-Challenge  & 93.10 & \textbf{93.13} & MBPP          & 71.84 & \textbf{72.97}\\
Logi. Deduct  & 80.81 & \textbf{81.03} & SciBench       & 28.33 & \textbf{29.64} & LiveCodeBench & 31.49 & \textbf{31.61}\\
AR-LSAT       & 37.00 & \textbf{38.00} & \cellcolor{black!10} & \cellcolor{black!10} & \cellcolor{black!10} & \cellcolor{black!10} & \cellcolor{black!10} & \cellcolor{black!10}\\
\hline
\textbf{Avg.} & 69.96 & \textbf{71.05} & \textbf{Avg.}  & 58.46 & \textbf{59.54} & \textbf{Avg.} & 63.34 & \textbf{63.91}\\
\bottomrule
\end{tabular}}
\end{table*}

The meta-awareness also benefits generalization ability of the reasoning model in out-of-domain logical, scientific, and coding benchmarks as shown in \Cref{tab:generalization}. For logical reasoning domain, we follow the setup of \citep{logicllm} and test on ProntoQA~\citep{prontoqa}, ProofWriter~\citep{proofwriter}, FOLIO~\citep{folio}, LogicalDeduction~\citep{bigbench}, and AR-LSAT~\citep{AR-LSAT}. For scientific reasoning, we use GPQA Diamond \citep{rein2024gpqa}, R-Bench \citep{Guo2025RBench}, ARC-Challenge \citep{Clark2018ARC}, and SciBench \citep{wang2024scibench}. For coding, we evaluate on EvalPlus \citep{liu2023your}, CRUX-O \citep{pmlr-v235-gu24c}, MBPP \citep{austin2021program}, and LiveCodeBench \citep{jain2025livecodebench}. Although \metaname{} is not explicitly trained for generalization, strengthening meta-awareness consistently enhances out-of-domain performance. The base model is Qwen3-14B-Base, with the same training and evaluation configurations stated in the experiments section.

\clearpage
\section{Meta Reward Code Snippet}\label{app:reward}
The implementation of our scoring mechanism is shown in the snippet below. We calculate a composite score based on the presence of mathematical notions, the length of the solution, and the difficulty pass rate.

\begin{pythonsnippet}{}
def compute_score(solution_str: dict, ground_truth: dict) -> float:
    
    # --- 1. Check Input Availability ---
    has_notion = "math_notion" in solution_str
    has_length = isinstance(solution_str.get("solution_length"), int)
    has_diff = isinstance(solution_str.get("pass_rate"), int)

    notion_score, length_score, acc_score = 0, 0, 0

    # --- 2. Calculate Notion Score ---
    if has_notion:
        pred_notions = solution_str["math_notion"]
        # Normalize to list
        if isinstance(pred_notions, str):
            pred_notions = [n.strip("[] ") for n in pred_notions.split(",")]
        
        # Filter notions already in problem text
        pred_notions = [n for n in pred_notions 
                        if n not in ground_truth["problem"]]
        
        # Count occurrences (Pos for correct resp, Neg for incorrect)
        notion_counts = {n: 0 for n in pred_notions}
        for resp, correct in zip(ground_truth["response"], ground_truth["score"]):
            for n in pred_notions:
                if n in resp:
                    notion_counts[n] += (1 if correct == 1 else -1)
                    
        # Score is ratio of notions with positive net utility
        scores = [1 if cnt > 0 else 0 for cnt in notion_counts.values()]
        if scores: 
            notion_score = sum(scores) / len(scores)

    # --- 3. Calculate Length Score ---
    if has_length:
        correct_lens = [l for l, s in zip(ground_truth["length"], 
                        ground_truth["score"]) if s == 1]
        p_len = solution_str['solution_length']
        if correct_lens:
            # Check if predicted length is within range of correct answers
            min_l, max_l = min(correct_lens), max(correct_lens)
            length_score = int(min_l < p_len < max_l)

    # --- 4. Calculate Accuracy Score ---
    if has_diff:
        avg_score = sum(ground_truth["score"]) / len(ground_truth["score"])
        pred_score = solution_str["pass_rate"] / 8.0 
        acc_score = (0.01) ** abs(avg_score - pred_score)

    return (notion_score + acc_score + length_score) / 3
\end{pythonsnippet}
\clearpage
\section{Length Prediction and Training Dynamics}\label{app:len_tendency}
\begin{figure}[t]
    \centering
    \includegraphics[width=0.7\linewidth]{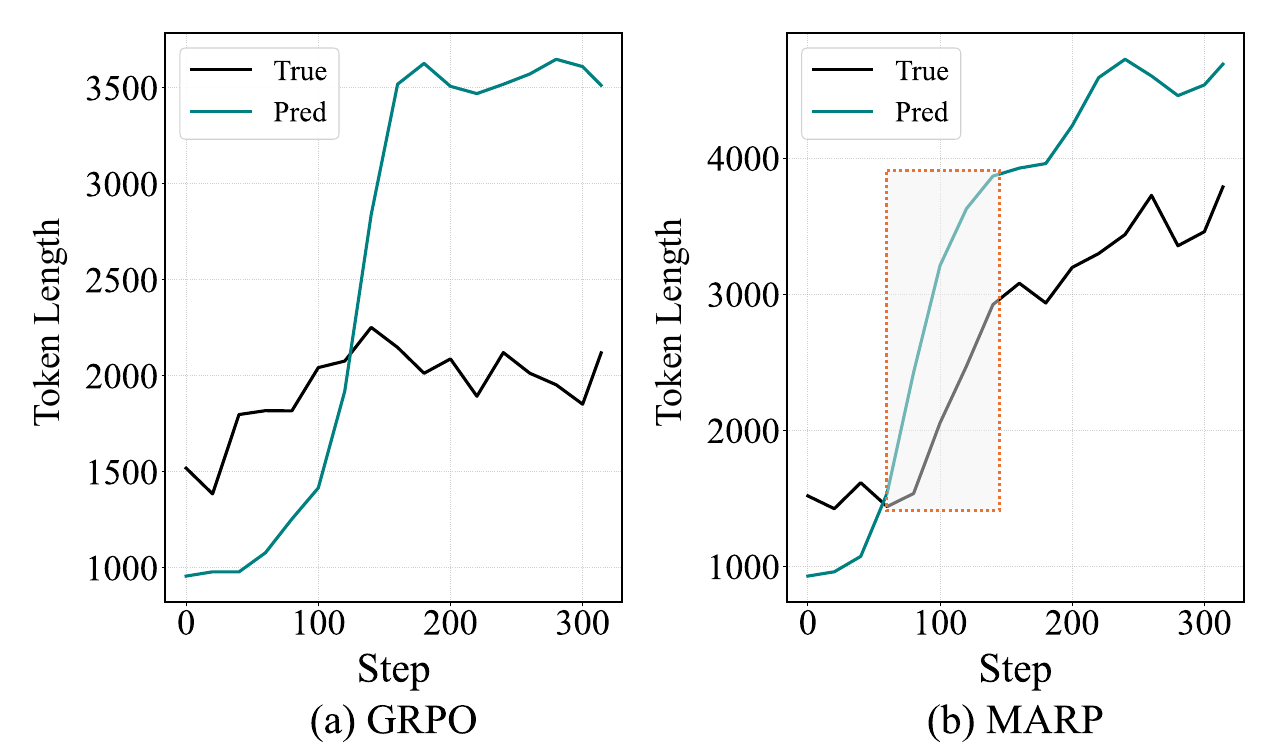}
     \caption{\textbf{}}
     \label{fig:length_dynamics}
\end{figure}
Similar to the observations on the difficulty prediction and training dynamics, we observe a surge in predicted length from the initial incorrect and underestimated state coincides with the rapid gain in performance during the training phase. This observation, coupled with the similar tendency in difficulty estimation, implies that calibration in the model's meta-awareness induces performance gain in reinforcement learning.

\clearpage
\section{Shapley $R^2$ Computation Details}
We first fit a linear model using all $p$ features to obtain the full-model coefficient of determination $R^2_{\text{full}}$. To compute feature-level contributions, we consider all permutations of the feature set. For each permutation, features are added sequentially to the model, and the marginal increase in $R^2$ upon adding feature $j$ is recorded. The Shapley contribution of feature $j$ is then defined as the average of its marginal $R^2$ gains over all permutations. This decomposition yields an additive attribution of $R^2_{\text{full}}$, providing a principled measure of each factor's explanatory power while accounting for feature interactions and ordering effects.

\section{Discussions}
\begin{table*}[h]
\centering
\caption{Examples of cross-domain notion prediction on coding and science tasks.}
\label{tab:cross_domain_notions}

\resizebox{0.95\linewidth}{!}{
\begin{tabular}{l|l|l}
\toprule
Domain & Task & Extracted Notions \\
\midrule

\multirow{3}{*}{Coding} 
& \texttt{maximum-strength-of-a-group} 
& array manipulation, mathematical operations, dynamic programming, greedy algorithms \\

& \texttt{find-the-longest-equal-subarray} 
& sliding window, hash map, two pointers \\

& \texttt{greatest-common-divisor-traversal} 
& graph traversal, GCD calculation, prime factorization \\

\midrule

\multirow{3}{*}{Science} 
& Quantum mechanics problem 
& quantum mechanics, Heisenberg uncertainty principle \\

& Organic chemistry synthesis 
& organic chemistry, Grignard reactions, oxidation, reaction mechanisms \\

& Gene interaction problem 
& epistasis, transcription factor, gene redundancy \\

\bottomrule
\end{tabular}
}
\end{table*}

Our notion-based reward formulation is not restricted to mathematical reasoning and can naturally generalize to other domains such as coding and science question answering. In coding tasks, notions correspond to high-level algorithmic concepts and data structure patterns, while in scientific reasoning they map to domain-specific scientific principles and terminology. To verify this, we apply a simple prompt adaptation without additional domain-specific training and analyze the generated notion predictions across domains. As shown in \Cref{tab:cross_domain_notions}, the model consistently extracts meaningful and task-relevant notions for both coding and science problems, suggesting that notion prediction captures transferable high-level semantic abstractions beyond mathematics.


\end{document}